\documentclass[acmsmall]{acmart}

\usepackage{xcolor,colortbl}
\AtBeginDocument{%
  }

\setcopyright{licensedothergov}
\acmYear{2024}
\acmVolume{7}
\acmNumber{1}
\acmArticle{1}
\acmMonth{5}
\acmDOI{10.1145/3651282}

\acmJournal{PACMCGIT}

\acmSubmissionID{1}

\usepackage{soul}
\usepackage{tikz}
\usetikzlibrary {decorations.fractals,spy}

\newcommand{\DMEM}{27}
\newcommand{\DPSNR}{0.21}

\newcommand{\resultfigwidth}{.193\linewidth}

\begin{document}

\title{Reducing the Memory Footprint of 3D Gaussian Splatting}

\author{Panagiotis Papantonakis}
\email{panagiotis.papantonakis@inria.fr}
\affiliation{
  \institution{Inria, Universit\'e C\^ote d'Azur}
  \country{France}
}

\author{Georgios Kopanas}
\email{george.kopanas@gmail.com}
\affiliation{
  \institution{Inria, Universit\'e C\^ote d'Azur}
  \country{France}
}

\author{Bernhard Kerbl}
\email{kerbl@cg.tuwien.ac.at}
\affiliation{
  \institution{Inria, Universit\'e C\^ote d'Azur}
  \country{France}
}
\affiliation{
	\institution{TU Wien}
	\country{Austria}
}

\author{Alexandre Lanvin}
\email{Alexandre.Lanvin@inria.fr}
\affiliation{
  \institution{Inria, Universit\'e C\^ote d'Azur}
  \country{France}
}

\author{George Drettakis}
\email{George.Drettakis@inria.fr}
\affiliation{
  \institution{Inria, Universit\'e C\^ote d'Azur}
  \country{France}
}


\begin{abstract}
3D Gaussian splatting provides excellent visual quality for novel view synthesis, with fast training and real-time rendering; unfortunately, the memory requirements of this method for storing and transmission are unreasonably high. We first analyze the reasons for this, identifying three main areas where storage can be reduced: the number of 3D Gaussian primitives used to represent a scene, the number of coefficients for the spherical harmonics used to represent directional radiance, and the precision required to store Gaussian primitive attributes. We present a solution to each of these issues. First, we propose an efficient, resolution-aware primitive pruning approach, reducing the primitive count by half. Second, we introduce an adaptive adjustment method to choose the number of coefficients used to represent directional radiance for each Gaussian primitive, and finally a codebook-based quantization method, together with a half-float representation for further memory reduction. Taken together, these three components result in a $\times$\DMEM\ reduction in overall size on disk on the standard datasets we tested, along with a $\times$1.7 speedup in rendering speed. We demonstrate our method on standard datasets and show how our solution results in significantly reduced download times when using the method on a mobile device (see Fig.~\ref{fig:webgl}).

\end{abstract}

%



\begin{teaserfigure}
	\centering
	\includegraphics[width=\textwidth]{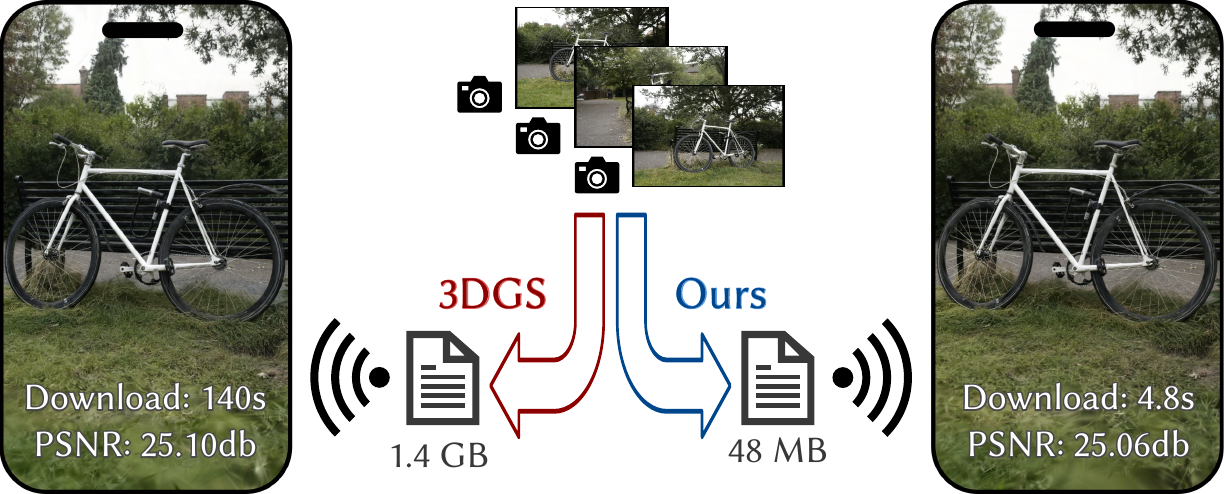}
	\caption{
		\label{fig:webgl}
		Left: screenshot from a phone running a modified \texttt{gsplat.js} for the bicycle scene. Right: same scene, processed with our method, reaching a significantly lower memory footprint and shorter download time.
	}
\end{teaserfigure}

\maketitle

\section{Introduction}
Novel view synthesis (NVS), i.e., generating new 3D views of a scene captured with photographs or video, has seen impressive progress over the last 25 years, ever since the first work on light fields and view-dependent texturing~\cite{levoy1996light,VDTM}. Traditional image-based rendering~\cite{buehler2001unstructured,shum2008image} progressively adopted deep learning~\cite{hedman2018deep,riegler2020free}, and in recent years led to \emph{neural rendering}~\cite{tewari2022advances,xie2022neural} that has dominated the field. 
Previous methods provide different tradeoffs between speed, visual quality, and memory, usually sacrificing at least one; we present a memory-efficient approach based on 3D Gaussian Splatting (3DGS) that results in an NVS method that excels in all three criteria.

Neural Radiance Fields (NeRFs) ~\cite{mildenhall2020nerf,mipnerf360} introduced a new \emph{implicit} scene representation based on a Multi-Layer Peceptron (MLP), that encodes volumetric density as a proxy for geometry and directional radiance. Rendering is performed with ray-marching. This solution provided unprecedented visual quality for NVS, but came at the cost of very expensive optimization to train the MLP, and slow rendering times. Several methods accelerated training and rendering, typically using spatial data structures (e.g.,~\cite{dvgo-cvpr2022,plenoxels}) or encodings such as hashing~\cite{mueller2022instant}, but sacrificed visual quality. More recently 3DGS proposed an \emph{explicit} representation using 3D Gaussian primitives in space, each carrying a set of attributes such as position, covariance (anisotropic scale and rotation), opacity, and Spherical Harmonic (SH) coefficients to represent directional radiance. 3DGS combines the performance benefits of all previous methods: fast training, high visual quality, and is the first to provide real-time rendering speed without quality degradation.
However, the original method results in a representation with an unreasonably high memory footprint (700Mb-1.2Gb for the scenes presented). This memory footprint is problematic for storage and processing, and also prohibitive for streaming to mobile devices. 

The original 3DGS method starts with a sparse set of 3D primitives and progressively densifies based on gradients. This is an effective strategy, but is wasteful. Similarly, the approach uses three SH bands for all primitives in the scene, even in cases where there are no view-dependent effects, where a single color value would suffice. Finally, most primitive attributes do not require very high accuracy or dynamic range, making them amenable to quantization.

We first analyze these three aspects, i.e., number of primitives, SH band utility, and quantization, and propose a complete and effective solution.
To reduce the number of primitives, we extend the existing pruning scheme by developing a resolution-aware \emph{redundancy} score using an efficient two-step algorithm that is used to eliminate approximately 60\% of the primitives compared to the baseline. We introduce an adaptive adjustment method for SH bands, where we reduce the number of bands for each primitive during training. We do this with a combination of contribution estimation and multi-view consistency evaluation to identify primitives without view-dependent appearance. Finally, we present a codebook quantization approach for some attributes and use half-floats to further reduce the stored size of the representation. Taken together, these three steps reduce the size of the 3DGS representation by \DMEM$\times$, resulting in a 1.7$\times$ increase in rendering speed.
In summary, we present three main contributions:
\begin{itemize}
	\item An efficient, resolution-aware primitive pruning approach, that reduces unnecessary points during optimization, leading to a total reduction of the primitive count by 60\%.
	\item An adaptive adjustment method to choose the number of SH bands required for each 3DGS primitive, significantly reducing overall memory footprint.
	\item A codebook-based quantization method, together with a half-float representation for more efficient storage of the representation, resulting in further memory reduction.
\end{itemize}
Even though our ultimate goal is to reduce the final, stored size of the representation,
depending on the implementation,
the resulting 3DGS model can have significantly reduced memory requirements at inference-time rendering.
We present a complete evaluation of the standard set of datasets used in 3DGS~\cite{3DGS}, showing the effect on memory reduction and rendering speed across different scene types. We also present an implementation of our method in a WebGL framework~\cite{dylanWebGL} that demonstrates 20--30 times faster downloads 
on a mobile phone.
\section{Related Work}
Research in Novel View Synthesis (NVS) from an unstructured set of input photographs has seen an explosion in recent years~\cite{tewari2022advances}. Throughout the various available solutions, we observe different compromises between quality, speed, and memory footprint.
In this section we briefly review the evolution of NVS algorithms, starting with Image-Based Rendering (IBR), all the way to the most recent real-time radiance field methods.

IBR algorithms typically use a geometry \emph{proxy}, usually a triangle mesh, to re-project information stored in the input views to the novel view. Popular implementations~\cite{sibr2020} of the Unstructured Lumigraph Rendering~\cite{buehler2001unstructured} use Multi-View Stereo to create the proxy geometry, reproject the RGB colors of the input views to the novel view and blend them using handcrafted heuristics to render~\cite{eisemann2008floating,chaurasia2013depth}. Such heuristics were later replaced by deep learning methods trained on multi-view datasets~\cite{hedman2018deep}. 
These techniques achieve satisfactory display speed but are prone to artifacts generated by reconstruction errors in the proxy geometry. Regarding memory, these methods need to store all input images on the GPU: this does not scale well as the complexity of the scene increases and we need more input views to represent it. 

Recently, Neural Radiance Fields (NeRFs)~\cite{mildenhall2020nerf} introduced a differentiable volumetric representation for NVS. The differentiable nature of the representation allows to optimize the properties of the scene with Stochastic Gradient Descent (SGD) to fit the input images. In the original paper, the scene is represented by a very compact, but slow Multi-Layer Perceptron (MLP).
The MLP is an \emph{implicit} representation, encoding radiance and density as a proxy for geometry; it is rendered with ray-marching, which requires the evaluation of the MLP at each sample.
Even for small scenes, using the MLP results in optimization (``training") times that take days and rendering that requires minutes per frame. Several improvements followed~\cite{barron2021mipnerf,mipnerf360}, but the underlying complexity remained. On the other hand, MLPs are an extremely compact representation that only needs a handful of megabytes (MB) of memory to enable rendering in high quality.

To improve on speed, both in terms of rendering and training, some methods~\cite{dvgo-cvpr2022} store the scene representation in a voxel grid, often paired with a shallow MLP~\cite{mueller2022instant} to decode latent features to density and color. To deal with the cubic memory complexity of voxel grids, various methods suggest sparse structures~\cite{liu2020neural,yu2021plenoctrees, plenoxels} or compression schemes using hash functions~\cite{mueller2022instant} and tensor decompositions~\cite{tensorf-eccv2022}. As a noteworthy example, Zip-NeRF~\cite{zipnerf} combines the benefits of discrete hash-based representations~\cite{mueller2022instant} with the anti-aliasing properties of MipNerf360~\cite{mipnerf360}. MeRF~\cite{merf} combines a low-resolution voxel grid and a tri-plane decomposition to achieve memory footprint compression of voxel-based representations in unbounded scenes. Shell-based methods~\cite{wang2023adaptiveshells} and adaptive surfaceness~\cite{turki2023hybridnerf} recently helped to accelerate rendering speed by using fewer samples per ray, but still suffer from limitations in quality and speed.
In all these cases, there is a significant trade-off between quality, speed, and memory consumption. When the extent of the scenes is relatively large, i.e., the MipNeRF360 dataset~\cite{mipnerf360}, this becomes more apparent: for a given memory footprint, the quality of grid-based solutions quickly deteriorates as the scene grows. The interplay between speed, quality, and memory consumption in real-world scenes is analyzed and presented in Sec.~\ref{sec:eval}. 

When real-world scenes are represented as a volumetric voxel grid, most of the space remains unoccupied.  
Many previous techniques use data structures or algorithms to skip empty space:
Instant-NGP~\cite{mueller2022instant} uses an occupancy grid during traversal, while  Plenoctrees~\cite{yu2021plenoctrees} prune an octree to remove nodes that represent empty space.
In contrast, 3D Gaussian Splatting (3DGS)~\cite{3DGS} is an inherently sparse representation for radiance fields, allowing the gradual formation of anisotropic 3D Gaussians---i.e., volumetric primitives in space amenable to rasterization---thus naturally omitting empty space.
The primitive-based nature of 3DGS avoids the cubic complexity of voxel-based implicit radiance field representations, 
while achieving quality that is usually on par (and sometimes surpasses) high-quality methods (e.g., MipNeRF360~\cite{mipnerf360}). Most importantly, 3DGS achieves unprecedented rendering speeds of 100+ frames per second (FPS) at this quality.
However, in the original paper, the memory footprint of the 3DGS representation is significantly higher than previous methods.
In this paper, we identify and address the factors that make 3DGS use excessive amounts of memory.

In unpublished, concurrent work (preprints), significant improvements have been made for voxel-based representations~\cite{duckworth2023smerf}. Similarly, several such preprints
propose methods to compress the 3DGS representation~\cite{lee2023compact,navaneet2023compact3d,fan2023lightgaussian,girish2023eagles}, by at least an order of magnitude. These works explore similar solutions to ours, but either focus on a single way to compress the scene~\cite{navaneet2023compact3d} or do not cull redundant Gaussians as we do~\cite{girish2023eagles}. The method closest to ours~\cite{fan2023lightgaussian} achieves 37\% lower compression than ours, with similar quality reduction. We provide a detailed comparison to concurrent work in the Appendix.
Outside academia, 3DGS has attracted the interest of practitioners, who have explored the potential to compress a 3DGS scene with hardware-accelerated texture compression of GPUs ~\cite{aras_compress}, which can be seen as a hardware-oriented alternative to our codebook compression.

\section{Analyzing 3DGS Memory Usage}
\label{sec:analysis}
Our goal is to significantly reduce the memory footprint of 3DGS, bringing it to the level of the most compact radiance field representation, while maintaining the speed and quality advantages of the initial method. 
To do this, we first 
analyze the memory footprint of the model and the various parameters that affect memory usage.
We start with a brief review of 3DGS, focusing on the parts that related to storage.

\begin{figure}
	\includegraphics[width=\textwidth]{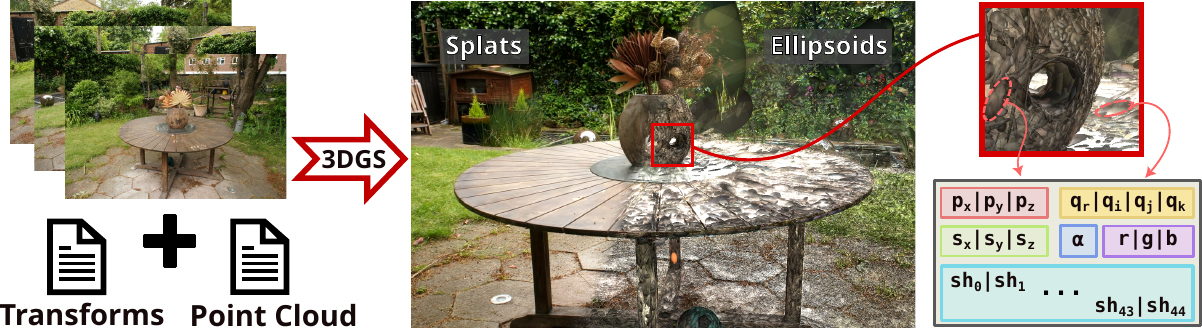}
	\caption{
		\label{fig:3DGS}
		3DGS produces photo-realistic renderings from input views and sparse points. Rendering the Gaussians as ellipsoids instead of splats reveals that each scene is modeled by millions of primitives. Each primitive stores a significant amount of information: position $p$, rotation quaternion $q$, scale $s$, opacity $\alpha$, color, and 3 bands of spherical harmonics. This leads to the exceedingly high memory consumption of 3DGS scenes.
	}
\end{figure}
3DGS models a scene as a set of 3D Gaussian primitives (which we represent as ellipsoids, see Fig.~\ref{fig:3DGS}, right), each centered at a given point. Each primitive has a set of attributes, namely opacity (used for alpha-blending), covariance (i.e., anisotropic scale and rotation), and spherical harmonics (SH) to represent diffuse directional radiance.
The technique is usually initialized with the point cloud produced by the camera calibration algorithm. 
This initial point cloud is far too sparse for novel view synthesis; consequently, 3DGS optimization includes a method for
adaptive control of the number of Gaussians.
In areas that are initially empty or missing details, the 3D Gaussians are densified by adding more primitives. A simple measure is used to control densification, namely the magnitude of the gradient of primitive positions: where the gradient is large, primitives are added.
In addition, primitives that have low opacity (and thus do not contribute in the rendering),
or primitives with a very large world space size, are regularly culled.

We first analyze the resulting representation, observing
that in many cases, 3DGS creates unnecessarily dense sets of primitives. 
An ideal densification strategy would avoid this in the first place; however, reliably identifying the density of primitives required is hard since it would amount to knowing the solution to the optimization problem beforehand. Consequently,
we address the issue of excessive primitive density by determining which primitives are \emph{redundant}; as we show later, this is strongly dependent on the scale and resolution of the observed details.

The memory footprint of the attributes for a given 3DGS primitive is as follows: position, scale, and color (3 floats each), rotation (4 floats for a quaternion), opacity (1 float), and the SH coefficients. 
This leads to a primitive structure of $14 + 3\sum_{i=1}^{N}(2i + 1)$ floats,
where $N = 0, 1, 2, \dots $ is the number of SH bands.
When using 3 bands as in the original paper, each primitive requires 59 floats of storage,
of which 45 (or $76\%$) are used by SH to model view-dependent effects.


Our second observation is that this results in
wasteful memory utilization, as most parts of the scenes include entirely or mostly diffuse materials,
which can in principle be modeled well with just the base (RGB) color.
In many cases, SH is used to represent view-dependent material appearance. It is however important to note that view-dependent effects in 3DGS and similar methods can also be modeled by combinations of several primitives. 
A typical case is the ``reflected geometry'' that such methods create behind a reflective surface, used to model moving highlights or reflections.
We introduce a new approach that uses
a \emph{variable number} of spherical harmonics bands, enabling them only when they are actually required, 
thus
allocating memory where it is really needed.

%

Our third observation relates to the \emph{effective dynamic range} and \emph{required accuracy} of many of the primitive attributes. Notably, opacity, scale, rotation, and the SH coefficients do not
require very high dynamic range, nor are they very sensitive to small inaccuracies. We will exploit this fact to introduce a post-processing step that further compresses the representation, using a clustering-based codebook approach. In contrast, reducing the accuracy of the \emph{positions} of the primitives in a similar fashion leads to a significant degradation in quality. 

These three observations lead us to the memory reduction methods we present next.

\section{Memory Reduction for 3DGS}

\begin{figure}[!h]
	\includegraphics[width=\textwidth]{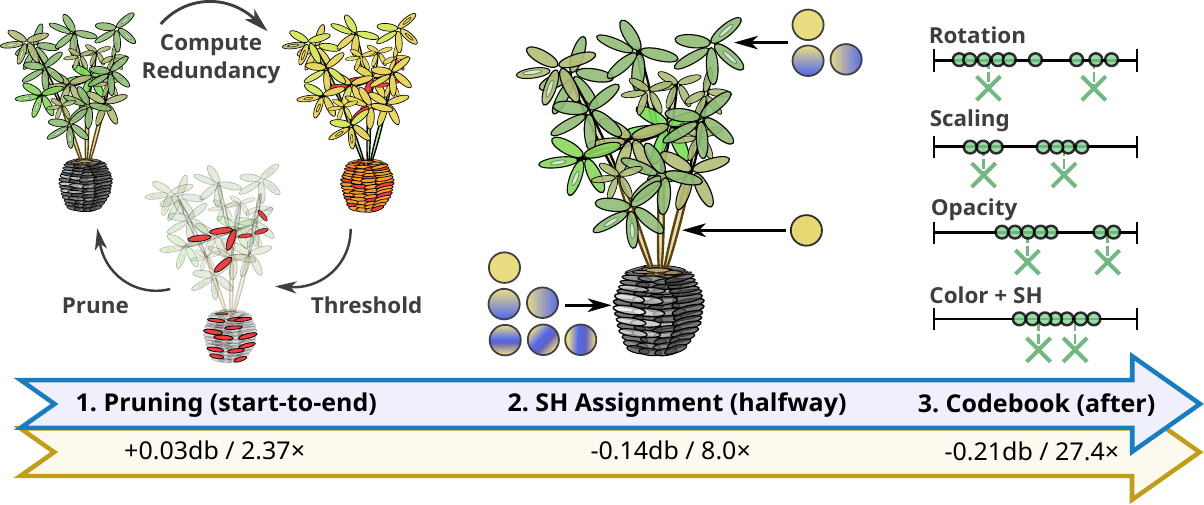}
	\vspace{-0.8cm}
	\caption{
		\label{fig:overview}
		Left: During training, every 1000 iterations our method evaluates a \emph{redundancy score} in space, which is then projected to the primitives. Redundant primitives are then culled. Middle: At 15K iterations when densification stops, our method analyzes the SH coefficients to determine which primitives can be represented with 0 (just RGB) 1, 2, or 3 SH bands, which allows us to omit storing unnecessary SH coefficients. Right: Finally, at the end of the training, we perform a codebook quantization of the remaining values, except for primitive positions. The relative reduction of each stage is shown in the figure, for a total of \DMEM~times reduction in memory with \DPSNR~db PSNR drop on average over all our datasets.
	}
\end{figure}

Based on the analysis presented above, we first present a method to identify redundant primitives and prune them \emph{during optimization}. Next, we introduce a method allowing a \emph{variable number} of SH bands for each primitive, and finally, a \emph{codebook compression} method that performs quantization as post-processing for attributes that do not require high accuracy.
An overview of our method is illustrated in Fig.~\ref{fig:overview}.

\subsection{Scale- and Resolution-aware Redundant Primitive Removal}
\label{sec:point-removal}


During optimization, the original 3DGS approach regularly culls primitives that fall below a specified opacity threshold,
as they contribute little to the final image.
To limit the number of primitives,
a straightforward solution would be to amplify this policy and simply remove low-opacity primitives more aggressively,
e.g., by erasing a specified percentage each time.
Doing so can already limit the final number of primitives, and hence the memory footprint of the representation.
However, as we have observed, 3DGS' adaptive densification can lead to regions with an exorbitantly high number of Gaussians.
Therefore, we propose to estimate this \emph{spatial redundancy} and combine this information with low-opacity filtering to achieve a highly effective culling strategy.

\begin{figure}
	\centering
	\includegraphics[width=1.0\linewidth]{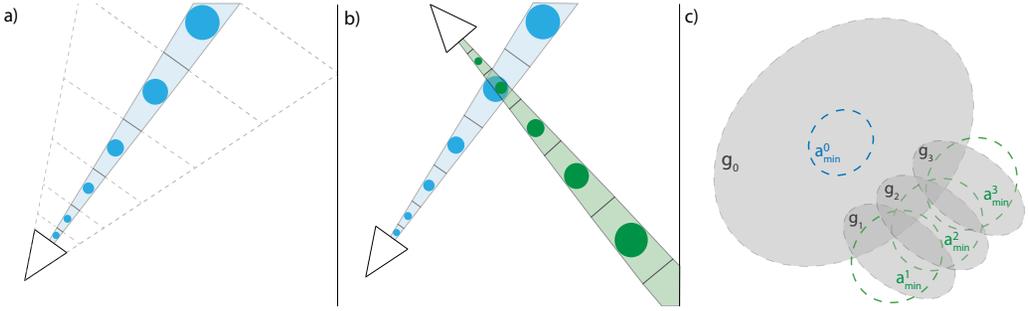}
	\caption{
		Our resolution and scale-aware redundancy metric measures how necessary a Gaussian is to represent the scene. a) Each camera can capture details of specific resolution, the further we move away from the camera, the smaller the spatial resolution this camera can represent. b) Given multiple cameras for a given primitive in the scene, multiple resolutions can be represented. c) For each Gaussian $g_i$, we consider the highest resolution $a_{min}^i$ given the input cameras. We count the number of Gaussians that intersect this region. Then we prune the Gaussians that intersect with regions that are influenced by more than K other Gaussians. In this example, Gaussian $g_0$ will \emph{not} be pruned because there is at least one region, $a_{min}^0$, influenced by no other Gaussian. While Gaussian $g_2$ intersects with regions $a_{min}^{1,2,3}$. All these regions have many Gaussians influencing them, hence $g_2$ is a good candidate for pruning.}
	\label{fig:pruning}
\end{figure}

The goal of our method is to first identify regions in space that have \emph{redundant} primitives. In essence, we are searching for regions where a large number of primitives exist, but here each of these primitives contributes little to the resulting rendered image. To make this decision for each Gaussian primitive $g$ in space, we find the number of other primitives $g'$ overlapping a spherical region around $g$.
To choose the sphere size, we take the world space \emph{footprint} of pixels that observe $g$ into account.
For a given view $j$, the pixel footprint at $g$ identifies the world-space extent of a pixel projected to $g$'s depth as seen from $j$ (see Fig.~\ref{fig:pruning}(a,b)).
Ideally, a 3D region around $g$ of that extent should be occupied by a low number of primitives; densely-packed clusters of Gaussians in that region would surpass the amount of detail that can be distinguished from $j$.

The sphere extent is chosen based on the pixel footprint of the closest view in which $g$ is visible (i.e., inside the camera frustum).
The closest view corresponds to the smallest pixel footprint $a_\textrm{min}$ at $g$, which yields a spatial extent proportional to the smallest observable detail around $g$ from any view (Fig.~\ref{fig:pruning}(c)).
This choice makes the redundancy test \emph{conservative}, i.e., we tend to \emph{underestimate} redundancy for most views.
In practice,
this means that our criterion is consistent with all novel views that observe a region at a smaller resolution (e.g. they are further away),
hence avoiding any visual degradation compared to the baseline.
For novel views that observe the region at a higher resolution,
our method retains the same level of information as the original solution.
This could be seen as taking the camera which determines the \emph{maximal sampling rate} of the primitive, following the terminology described in \cite{Yu2023MipSplatting}.
We then find the number of other Gaussian primitives intersecting a sphere around $g$ with a radius equal to $a_\textrm{min}\frac{\sqrt{3}}{2}$, i.e., half the diagonal of a cube where each face has area $a_\textrm{min}$.
For the purpose of this intersection test, each Gaussian is represented by an ellipsoid with its center at the Gaussian's mean and axis lengths/orientation corresponding to the Gaussian's scale/rotation, as defined by 3DGS.

Na\"ively computing the above redundancy score is prohibitively expensive for millions of Gaussians. 
To limit the number of intersection tests, we first apply a k-NN search with a high number of neighbors (30) on the set of primitives to find candidate intersecting primitives. In addition, we approximate the sphere-ellipsoid intersection test by scaling the ellipsoids' axes by the sphere radius and performing an ellipsoid-point intersection test, which is just a dot product.
We count the number of primitives that pass the intersection test, yielding the spatial redundancy value for the spherical region centered around each Gaussian. 

Counting the number of intersections can be seen as sampling a \emph{spatial redundancy score field} at points of interest; rather than sampling the field uniformly, we use the Gaussians' centers as sample locations and then propagate the scores computed at these locations to close-by primitives.
To ensure that our approach remains conservative, propagation of the redundancy score to a primitive is done by choosing the smallest score from all spherical regions where each ellipsoid intersects.
The motivation for this is straightforward: if a Gaussian overlaps a region that is not redundantly populated, it may contribute crucial detail there, which should be reflected by a low redundancy score.
We do this by keeping a mask of the intersection tests from the first step and taking the minimum value from the regions intersected, for every primitive. 

Each primitive now has an assigned redundancy score; We sort our primitives based on their score and filter those whose score is greater than an adaptive threshold $\tau_p~=~(\mu + \lambda_r\sigma)$. Here, $\mu$ and $\sigma$ are the mean and standard deviation of the redundancy score for all primitives. Effectively, this adaptive threshold prunes Gaussians whose redundancy score is more than $\lambda_r$ standard deviations away from the mean. We use $\lambda_r=1$ in all our experiments.

Since the redundancy score of each primitive is not independent of other primitives, deleting all filtered primitives with values greater than $\tau_p$ could lead to excessive primitive culling in some regions. Instead, we delete 50\% of filtered primitives. We choose primitives to delete based on low opacity, as opacity is an objective/independent estimator of the Gaussian's contribution to the visual result; culling primitives with low opacity has little impact on image quality.

We found that culling primitives with a redundancy score less than 3---i.e., a primitive contributes to at least one region where it coincides with just two other primitives---adversely affects quality. We thus modify $\tau_p$ to be $\tau_p~=~\textrm{max}~(\mu + \lambda_r\sigma, 3)$. In practice, this means that we do not delete primitives with a redundancy score lower than 4, since scores are integers.

We also modify the loss during training by adding an $L_1$ sparsity term on opacity to encourage opacity values to be as low as possible. 
This helps to encourage the creation of lower-contribution primitives and is particularly effective when we cull the bottom 50\% of redundant candidate primitives.
The culling process is applied every 1000 iterations during optimization. 

The results of this process are evaluated in detail in Sec.~\ref{sec:eval};
Interestingly, simple low-opacity culling gives results close to those achieved by redundancy culling.
However,
the performance of each is different depending on the dataset tested.
Instead,
we combine both, which results in consistently better performance in all cases,
with around 60\% of the primitives being culled, 
and minimal effect on visual quality.
For the simple low-opacity portion,
we remove 3\% of lowest opacity primitives each time,
with a maximum opacity threshold of 0.05,
as removing primitives with larger opacity values degraded the quality.

\subsection{Adaptive Adjustment of Spherical Harmonics Bands}

At a high level, spherical harmonics help represent parts of the scene with view-dependent effects, such as glossy/specular highlights, even though they also provide additional flexibility in the optimization.  
As discussed previously, the SH coefficients represent the majority of the memory footprint of the 3D Gaussians.


Our motivation is thus to only use as many SH bands as are required by a given Gaussian primitive; in many cases, a single RGB color value may suffice. To find the appropriate representation, we use a purely numerical metric to determine if a lower-order SH is sufficient to represent the radiance of the primitive. We do this by evaluating the SH function from all input views; This operation determines whether the primitive needs to represent view-dependent effects, i.e., if the fully-evaluated color changes significantly between views.
We then use this information to cull SH bands following two approaches.
The first is based on the observation that a primitive that receives the same color from all input views
does not need view-dependent effects;
the second is that if the color of a primitive does not change much when evaluated with fewer, lower band SH coefficients,
the higher-order ones can be ignored.

Concretely,
for the first approach,
we determine if
the evaluated color of a primitive from all viewpoints has low variance.
If this is the case, the Gaussian can be modeled with just a RGB color ($0^{th}$ band), and does not
need to represent view-dependent appearance (higher bands).
Specifically, for every viewpoint, we calculate the per-channel average $\mu_c$ and standard deviation $\sigma_c$ of the color $\mathbf{c_i}$ for each primitive, weighted by average transmittance:
\begin{eqnarray}
	\mathbf{\mu_c} = \frac{\sum_i^N\mathbf{c_i}\bar{T_i}}{\sum_i^N \bar{T_i}}, \hspace{1cm}
	\mathbf{\sigma_c} = \frac{\sum_i^N(\mathbf{c_i}-\mathbf{\mu_c})^2\bar{T_i}}{\sum_i^N \bar{T_i}},
\end{eqnarray}
where $N$ is the number of views containing the primitive,
$\bar{T_i}$ is the average transmittance of the primitive on the pixels splatted by view $i$, given by:
\begin{equation}
	\bar{T_i} = \frac{\sum_k^P T_{ik}}{P}, 
\end{equation}
where $P$ is the number of pixels that the primitive is splatted to in view $i$, and $T_{ik}$ is the transmittance of the primitive for a specific pixel $k$.
We then replace the RGB color of all primitives that have a $\sigma_c$ lower than a threshold $\epsilon_\sigma$ with the average color $\mu_c$ and disable all of their higher-order SH bands. 
$\epsilon_\sigma$ is a hyperparameter that we set to 0.04 in all our experiments.

For the second approach,
we identify primitives for which dropping the higher-order SH bands creates only a minor change in their evaluated color across all views. 
In these cases,
the color computation can use the smaller number of bands and ignore the SH coefficients belonging to the higher bands.
For every viewpoint,
we evaluate the color using only the SH coefficients belonging to the first $q$ SH bands for each primitive,
resulting in colors $c_{q}$, with $q\in[0,3]$
Then, we calculate the Euclidean distance between the full color $c_3$ and the remaining three,
resulting in three color distances $d_0, d_1, d_2$.
Finally, we take the average of each distance value across all views, weighted by average transmittance.
We finally choose the lowest band $q$ for which $d_q$ is below a threshold $\epsilon_\text{cdist}$ and remove the remaining higher bands; $\epsilon_\text{cdist}$ is a hyperparameter, set to 0.04.
After applying these steps,
some primitives maintain all initial bands,
while most end up with 2 or fewer SH bands, removing the need to store respective, higher-order SH coefficients.
For our main model,
the point distribution over bands was 89\%, 0.1\%, 2.7\%, 8.2\% for 0, 1, 2, 3 bands respectively.

Disabling SH bands as described can result in a small quality degradation; to mitigate this effect, we apply the SH culling once at the halfway point of 3DGS training (15K iterations, i.e., when densification stops) and allow the remaining optimization steps to compensate for the adjustment.
To complement our band reduction procedure, we again modify the original 3DGS optimization by introducing a sparsity loss on SH coefficients. Doing so discourages the use of higher bands, except where absolutely necessary.
As a result, disabling higher bands has a lower chance of adversely impacting image quality. 

We evaluate the effects of primitive reduction and adaptive SH adjustments in Sec.~\ref{sec:eval}; Taken together, after both processes have been applied, the memory reduction is approximately 87\%.

\subsection{Quantization of the Final Representation}

We now exploit our third observation (Sec.~\ref{sec:analysis}), that only limited dynamic range and precision need to be stored for most primitive attributes.

We create a codebook using K-means clustering, i.e., 
instead of storing the exact value of a property for every primitive,
we store an index to the closest value in a fixed-size codebook.
For vector attributes (e.g., scale), we maintain one shared codebook, but treat the three scalar components separately, leading to one index per component.
Our experiments show that 1-byte indices allow maximum compression with minimal quality degradation, and thus a codebook size of 256 entries.
E.g., if the initial cost for storing $N$ Gaussian scales (3D vectors containing floats) is $3N\times4$ bytes, the cost becomes $3N + 4\times256$ bytes, including the size of the shared codebook.

We create the following codebooks for Gaussian primitive attributes:
one for opacity, one for the three scaling components, one for the real part and one for the imaginary components of quaternion rotation, one for the coefficients of the base color, and one per 3-channel color component for each of the 15 SH coefficient groups.
Hence,
with this procedure,
we reduce the required memory for these attributes from $56N\times4$ bytes to $56N + 20\times256\times4$ bytes.
Since $N$ is a number in the hundreds of thousands or even millions,
the second, constant term that represents the overhead for the codebooks
is negligible.

Finally,
we found that applying 16-bit half-float quantization to the remaining, uncompressed floating point values (i.e., position and codebook entries), does not significantly affect the quality. Thus, we also employ this quantization to reduce our memory requirements even further.
In total, the average reduction in memory is 96.3\%, or almost \DMEM$\times$ compared to the original 3DGS file size. Again, we analyze the effect of each of our decisions in Sec.~\ref{sec:eval}.
\section{Implementation}

We implemented our method on top of the original, open-source 3DGS implementation.
We will release the source code of our own implementation\footnote{https://repo-sam.inria.fr/fungraph/reduced\_3dgs/} upon acceptance.

We modified the original method's simple CUDA k-NN routine to enable the identification of the nearest neighbor primitive IDs.
In contrast to the original approach, we continue culling low-opacity Gaussians (opacity < $\frac{1}{255}$) after the 15K iteration mark, since they are neither rendered nor optimized and thus provide no contribution to image quality.

To incorporate the variable number of SH coefficients with minimal changes to the file format, in practice, we store 4 sets of primitives, one for each number of SH bands (0, 1, 2, and 3). This has minimal effect on the parsing of the file.

Although the resolution-aware primitive pruning method and the adaptive adjustment for the SH bands
occur during training and affect GPU memory usage,
their effect becomes relevant after the point of peak usage,
so the memory requirements of training remain unaltered.
During rendering,
the model benefits from the reduced memory footprint 
caused by the smaller number of Gaussians and fewer dynamic SH bands.
Fetching a variable number of SH coefficients requires only minor modifications to the program logic, which incur no discernible slowdown.

We assess the effectiveness of our method in terms of its impact on file size.
In particular, one of the most useful applications is streaming 3DGS representations over the network. To showcase the benefits of our compression in a real-world use case, we have extended the open-source WebGL implementation of~\cite{dylanWebGL} to incorporate our modified representation. The original codebase did not include SH coefficient evaluation; we added it to visually reproduce the original 3DGS results and added display of our adaptive SH coefficient representation. We show results in Sec.~\ref{sec:results} and in the supplemental video.
In our current implementation,
the quantized representation is decompressed at parsing time rather than on-the-fly.
Furthermore, we do not yet employ codebooks during rendering. As demonstrated by concurrent work~\cite{westermann}, we expect that integrating these directly into the rendering pipeline would lead to even higher frame rates due to reduced memory traffic.
\section{Results and Evaluation}

We present extensive results and evaluation of our methods on the standard set of scenes used in the original 3DGS paper~\cite{3DGS}, namely the full MipNeRF360~\cite{mipnerf360} dataset, two scenes from Deep Blending~\cite{hedman2018deep} and two from Tanks\&Temples~\cite{knapitsch2017tanks}.
All results were obtained using an NVIDIA RTX A6000 GPU, running on Linux.
Please note that the FPS measurements of 3DGS variants on our system isolate the runtime of the CUDA rasterizer routine only, thus focussing on the speedup of image synthesis itself, excluding any graphics API overheads such as blitting or swapchain presentation.

\subsection{Results}
\label{sec:results}

In Tab.~\ref{tab:main-results}, we show the effectiveness of our solution. Our primitive reduction reduces the memory footprint to between 32\% to 52\% of the original method~\footnote{We ran all the tests using the codebase of the original method on Github that shows slightly improved numbers compared to the published paper; see also Tab.~\ref{tab:comparisons} below.}. The average impact on PSNR is minimal, between -.32 and +.16 dB,
that has minimal effect on visual quality. The reason why we not only see minimal degradation in the visual quality but in some cases improvements is that the primitive and SH culling can also act as a regularization strategy, forcing the optimization to find solutions that generalize better. We show that each one of our main contributions---the primitive and SH culling, and the quantization---are all contributing significantly and comparably to compressing the 3DGS representation. Each one of these elements approximately cuts down the size of the representation 3$\times$ to result in an average of \DMEM$\times$ reduction in size across all datasets used in 3DGS~\cite{3DGS}. We note that integrating codebooks and half-float type handling into the renderer would enable the same reduction for a scene's VRAM consumption as for its required disk storage.

\begin{table}[!htb]	
	\small
	\caption{
		\label{tab:main-results}
		We show the effect of each component of our method as we add them progressively. From top to bottom, we show the effect of the reduction of the number of primitives, adaptive SH adjustment, and quantization. In each case, we show the effect on the three test datasets, in terms of SSIM, PSNR, LPIPS, total memory size, and memory reduction. 
	}
	\scalebox{0.72}{
		\begin{tabular}{l|m{0.6cm}m{0.6cm}m{0.7cm}r|m{0.6cm}m{0.6cm}m{0.7cm}r|m{0.6cm}m{0.6cm}m{0.7cm}r}
			Dataset & \multicolumn{4}{c|}{Mip-NeRF360} & \multicolumn{4}{c|}{Tanks\&Temples} & \multicolumn{4}{c}{Deep Blending} \\
			Method/Metric &        $SSIM^\uparrow$ &  $PSNR^\uparrow$ & $LPIPS^\downarrow$ & Mem ($\times$Gain) &          $SSIM^\uparrow$ &  $PSNR^\uparrow$ & $LPIPS^\downarrow$ & Mem ($\times$Gain) &          $SSIM^\uparrow$ &  $PSNR^\uparrow$ & $LPIPS^\downarrow$ & Mem ($\times$Gain) \\
			\midrule
			Baseline       &       0.813 & 27.42 & 0.217 &        744MB  ($\times$1.0) &         0.844 & 23.66 & 0.178 &        412MB  ($\times$1.0) &         0.899 & 29.47 & 0.247 &        630MB  ($\times$1.0) \\
			+Point Culling &       0.814 & 27.43 & 0.220 &        339MB  ($\times$2.2) &         0.844 & 23.69 & 0.182 &        154MB  ($\times$2.6) &         0.902 & 29.57 & 0.247 &        224MB  ($\times$2.9) \\
			+SH Culling    &       0.811 & 27.18 & 0.225 &        102MB  ($\times$7.5) &         0.841 & 23.62 & 0.187 &         49MB  ($\times$8.0) &         0.903 & 29.67 & 0.248 &         62MB ($\times$10.3) \\
			+Quantisation  &       0.809 & 27.10 & 0.226 &         29MB ($\times$25.7) &         0.840 & 23.57 & 0.188 &         14MB ($\times$27.6) &         0.902 & 29.63 & 0.249 &         18MB ($\times$34.8) \\
		\end{tabular}
	}
\end{table}

\begin{figure}[!htb]
	\setlength{\tabcolsep}{1pt}
	\begin{tabular}{ccccc}
		A & B & Ours & Error Image & Baseline \\
		\includegraphics[width=\resultfigwidth]{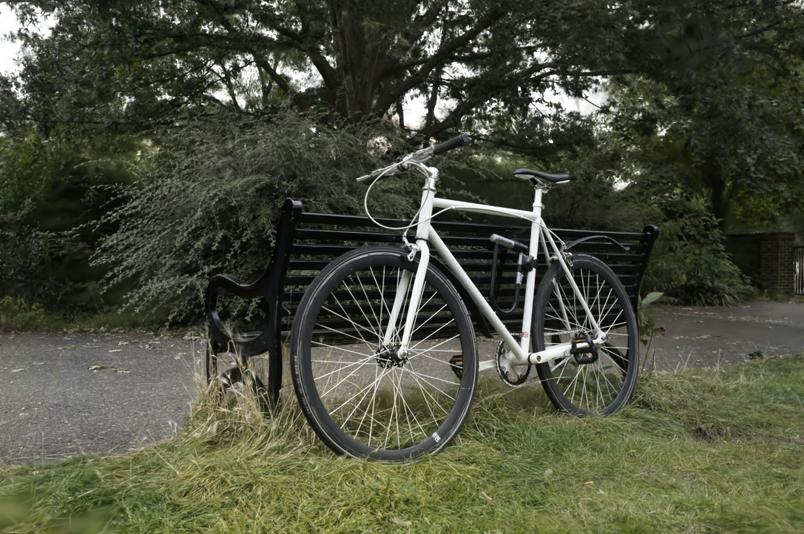} &
		\includegraphics[width=\resultfigwidth]{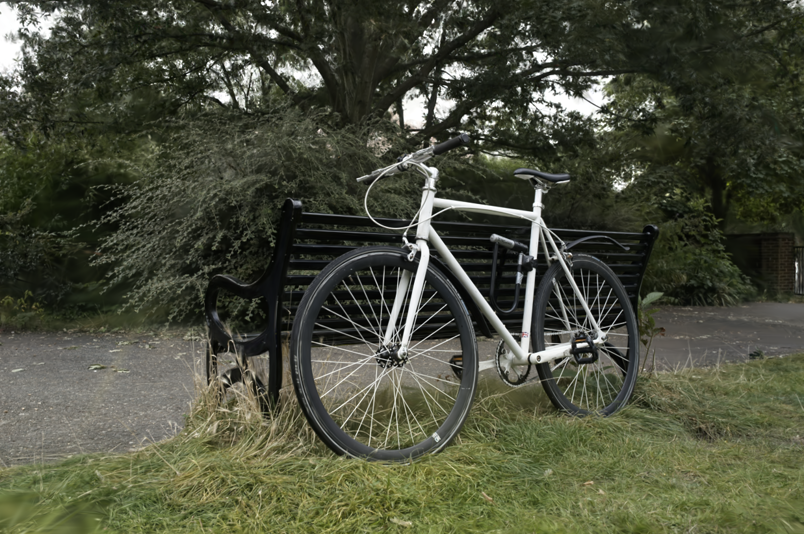} &
		\includegraphics[width=\resultfigwidth]{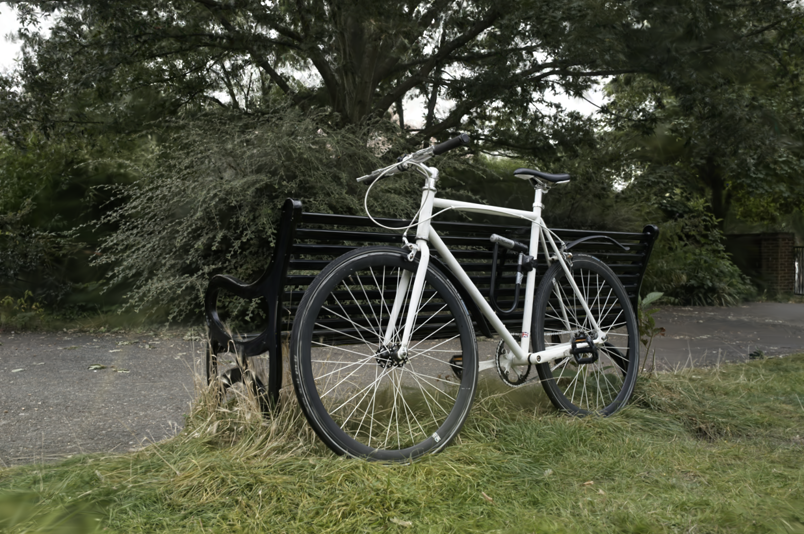} &
		\includegraphics[width=\resultfigwidth]{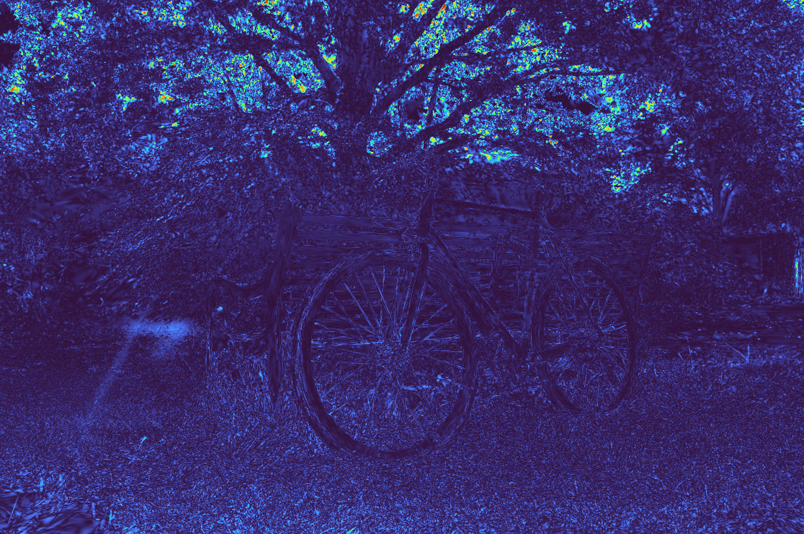} &
		\includegraphics[width=\resultfigwidth]{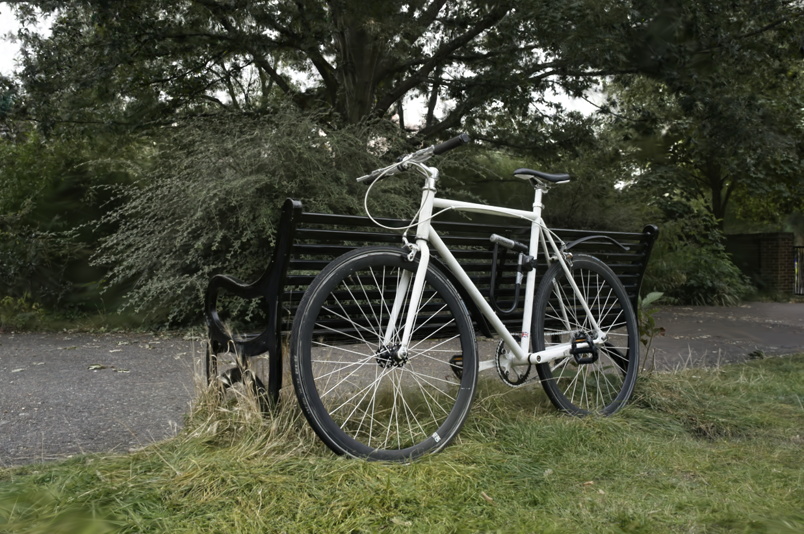} \\
		
		\includegraphics[width=\resultfigwidth]{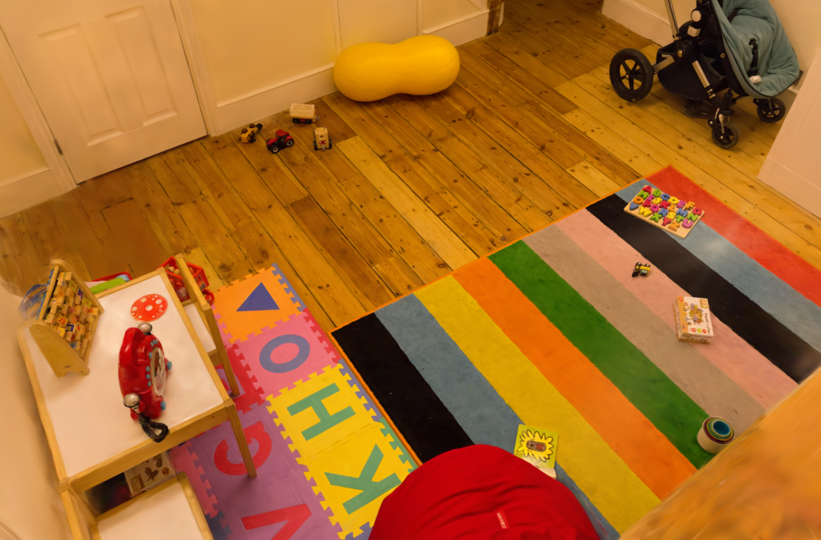} &
		\includegraphics[width=\resultfigwidth]{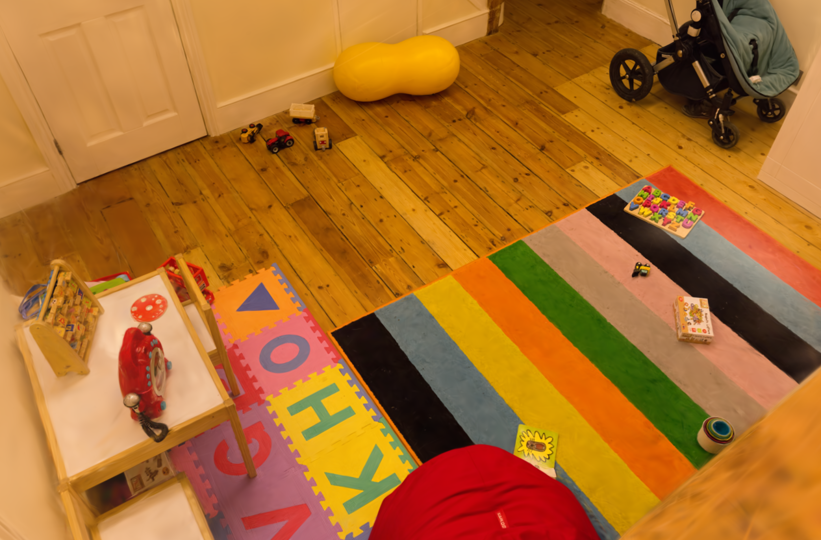} &
		\includegraphics[width=\resultfigwidth]{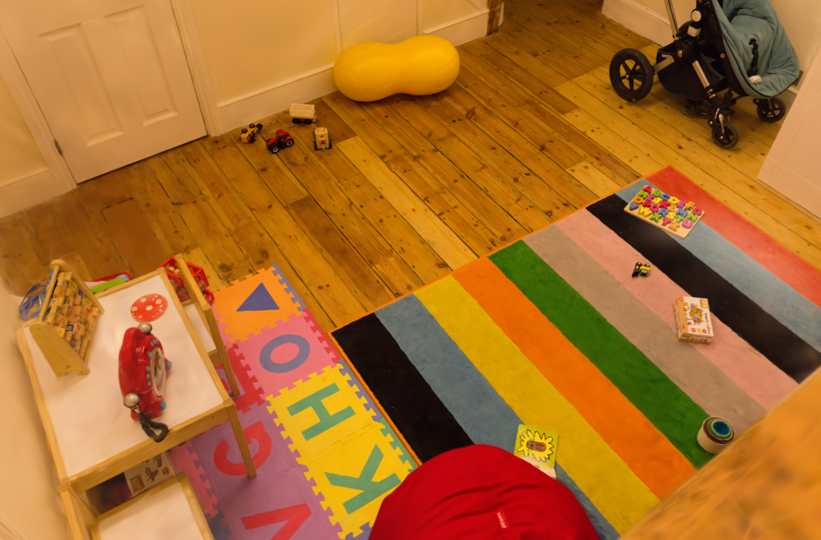} &
		\includegraphics[width=\resultfigwidth]{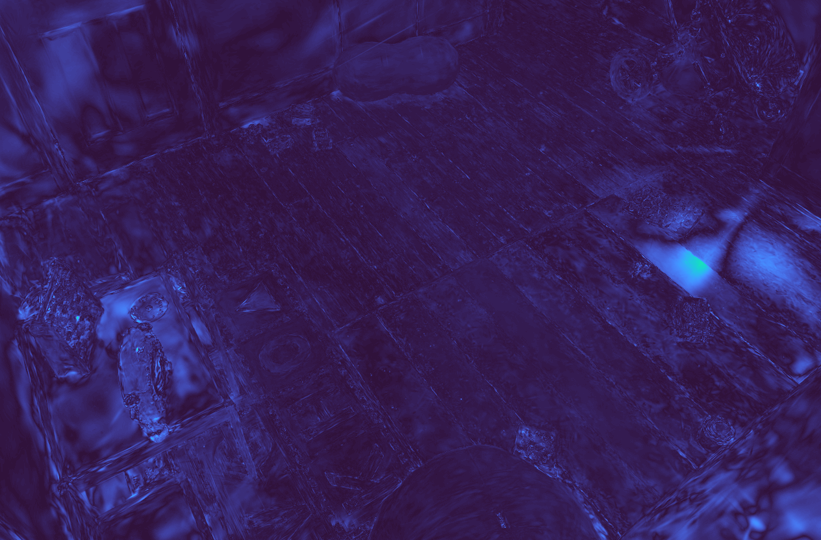} &
		\includegraphics[width=\resultfigwidth]{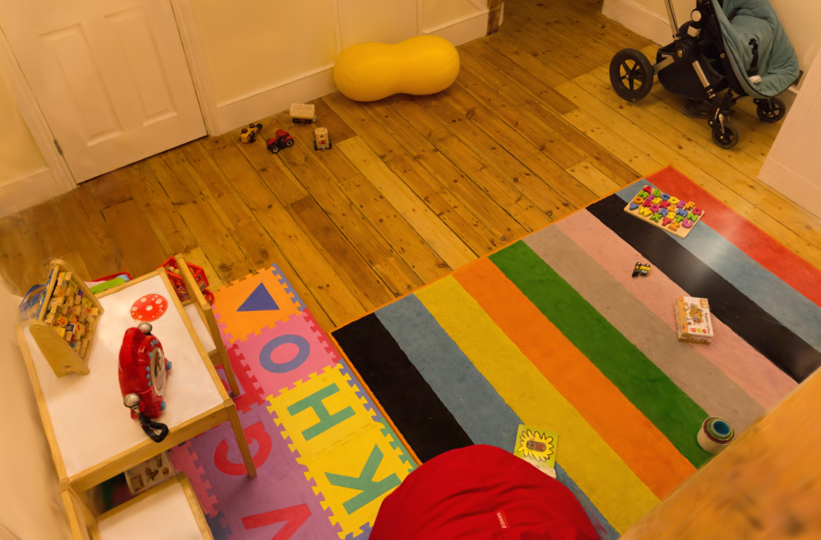} \\
		
		\includegraphics[width=\resultfigwidth]{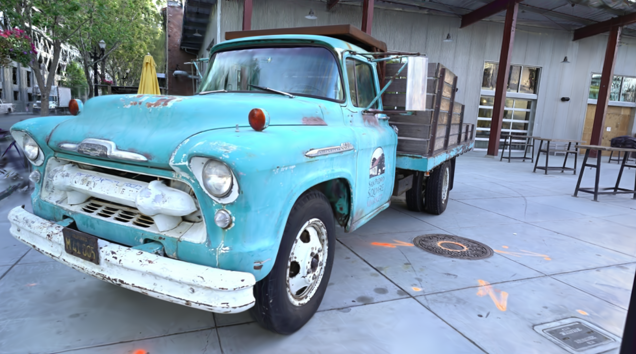} &
		\includegraphics[width=\resultfigwidth]{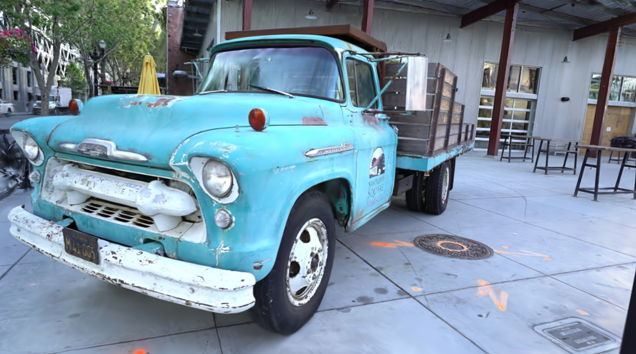} &
		\includegraphics[width=\resultfigwidth]{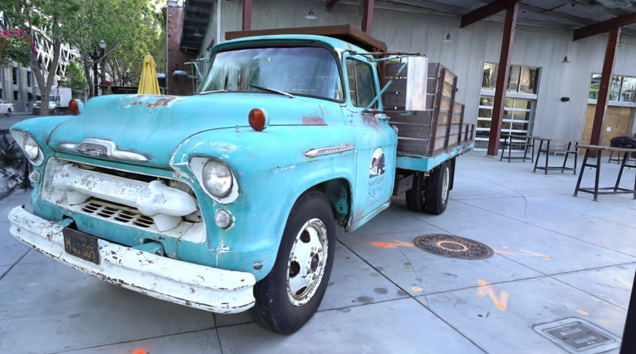} &
		\includegraphics[width=\resultfigwidth]{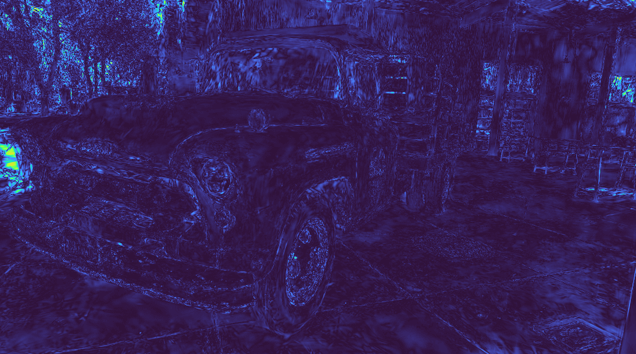} &
		\includegraphics[width=\resultfigwidth]{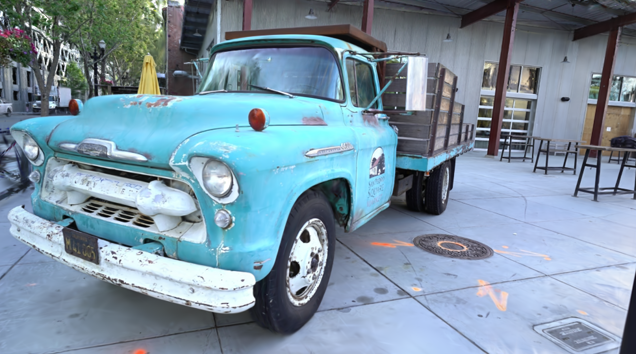} \\
		
	\end{tabular}
	\caption{
		\label{fig:res1}
		From Left to Right: Primitive Reduction only (A), Primitive Reduction and Adaptive SH (B), full method, error image between ours and baseline, and the baseline (original 3DGS). We show the scenes: Bicycle from MipNeRF360, Playroom from Deep Blending, and Truck from Tanks\&Temples.
	}
\end{figure}

This makes 3DGS representations more practical for applications, by making streaming assets through the web faster and loading them quicker. Our modifications incur a virtually imperceptible degradation in visual quality, especially for smaller displays. In Fig.~\ref{fig:res1}, we show the visual effect of each step for three scenes, one from each tested dataset. We show a test input view to allow comparison with ground truth. In print, the difference is invisible; the visual quality degradation is minimal, even when viewed on a computer screen. This is in contrast to other state-of-the-art real-time methods, where quality differences are more apparent. See Fig.~\ref{fig:res2} for an illustration.

\newcommand{\spyimage}[4]{%
	\begin{tikzpicture}[spy using outlines={circle,yellow,magnification=5,size=1cm, connect spies}]
		\node[anchor=south west,inner sep=0]  at (0,0) {\includegraphics[width=#1]{#2}};
		\spy on (#3) in node [left] at (#4);
	\end{tikzpicture}%
}

\begin{figure}[!htb]
	\setlength{\tabcolsep}{1pt}
	\begin{tabular}{ccccc}
		Ground Truth & 3DGS & Ours & INGP & MeRF \\
		
		\spyimage{\resultfigwidth}{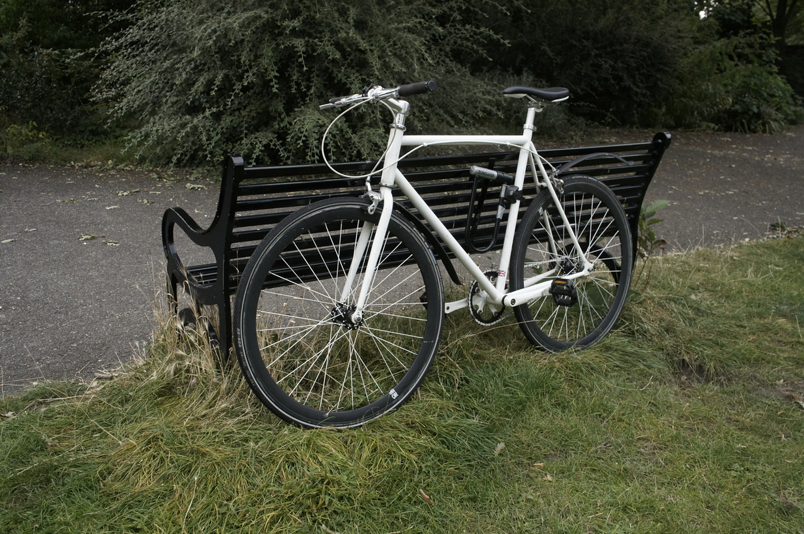}{1.2,0.7}{1,0.5} &
		\spyimage{\resultfigwidth}{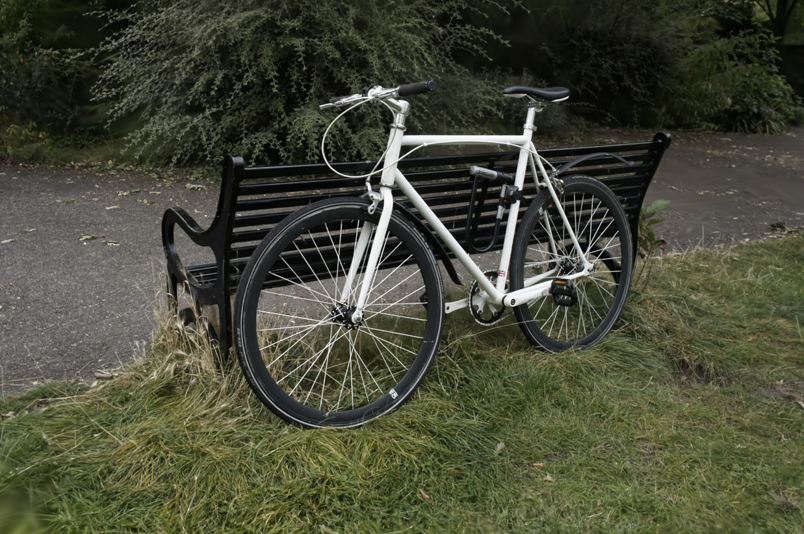}{1.2,0.7}{1,0.5} &
		\spyimage{\resultfigwidth}{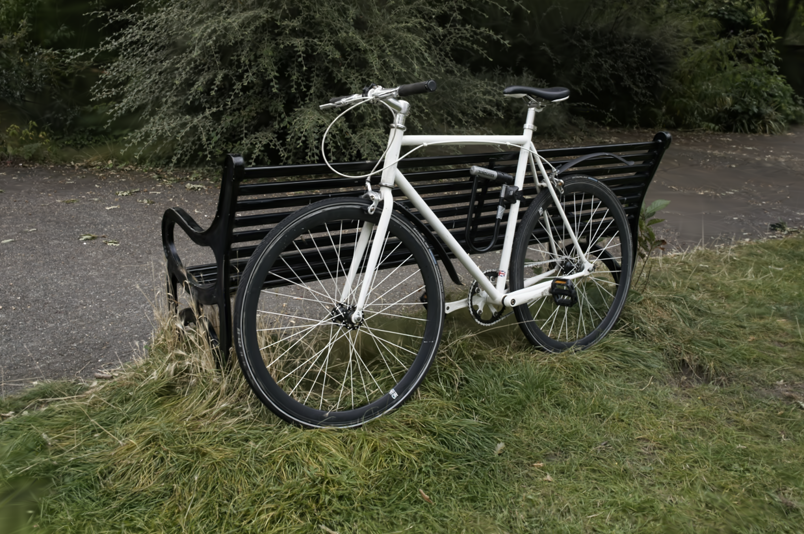}{1.2,0.7}{1,0.5} &
		\spyimage{\resultfigwidth}{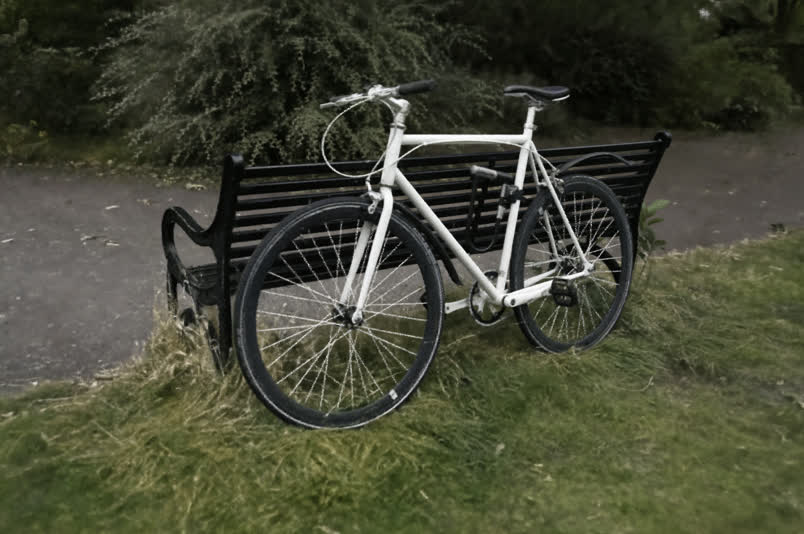}{1.2,0.7}{1,0.5} &
		\spyimage{\resultfigwidth}{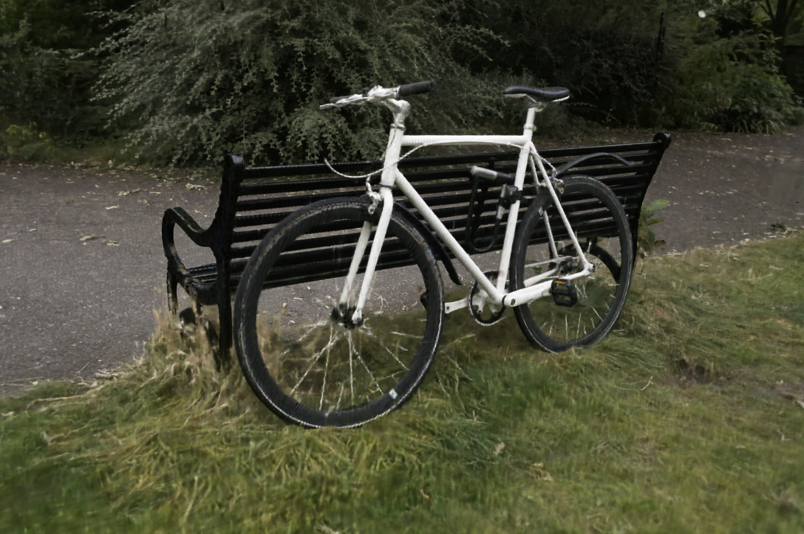}{1.2,0.7}{1,0.5} \\

		\spyimage{\resultfigwidth}{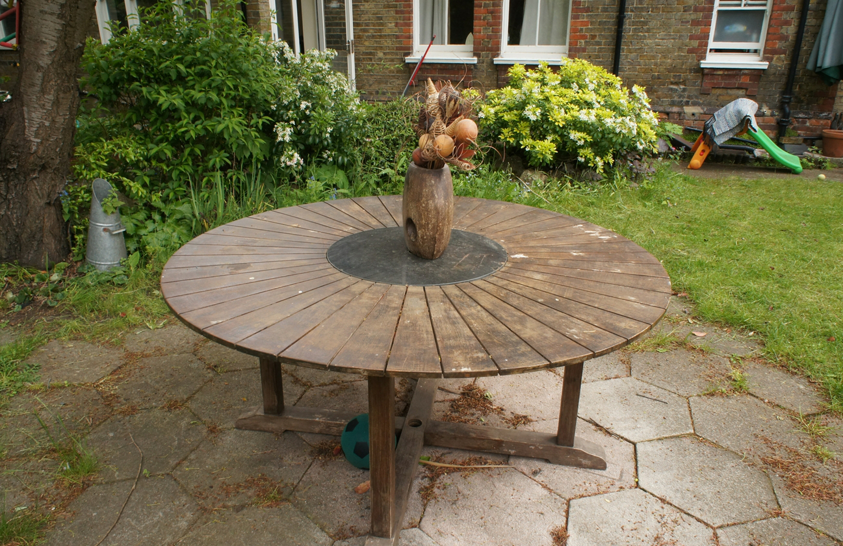}{1.4,1.4}{2.5,0.5} &
		\spyimage{\resultfigwidth}{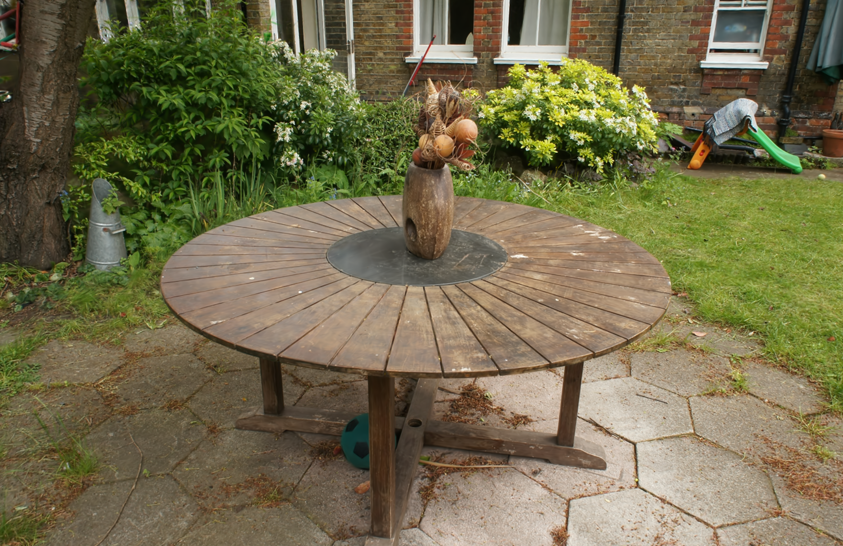}{1.4,1.4}{2.5,0.5} &
		\spyimage{\resultfigwidth}{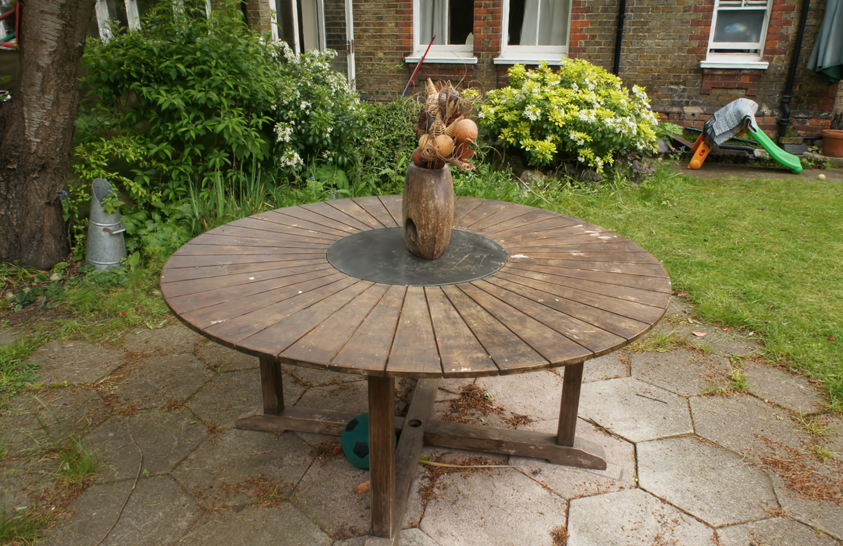}{1.4,1.4}{2.5,0.5} &
		\spyimage{\resultfigwidth}{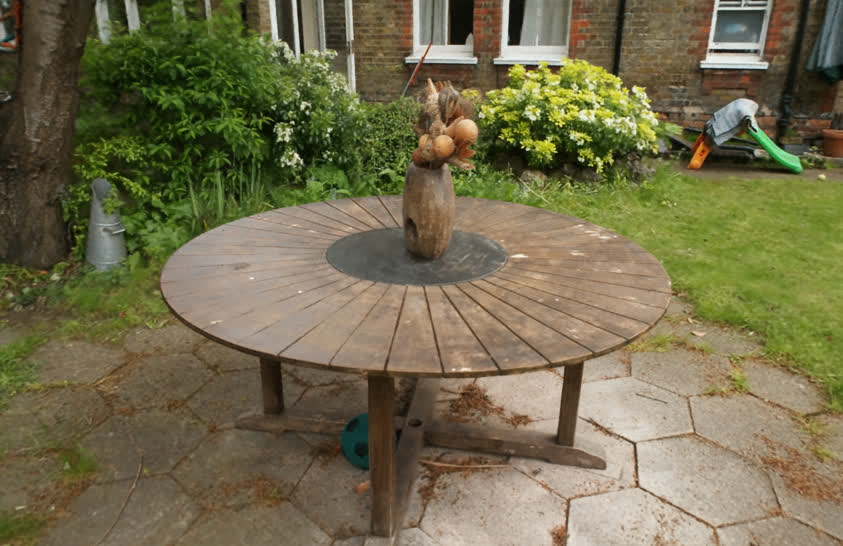}{1.4,1.4}{2.5,0.5} &
		\spyimage{\resultfigwidth}{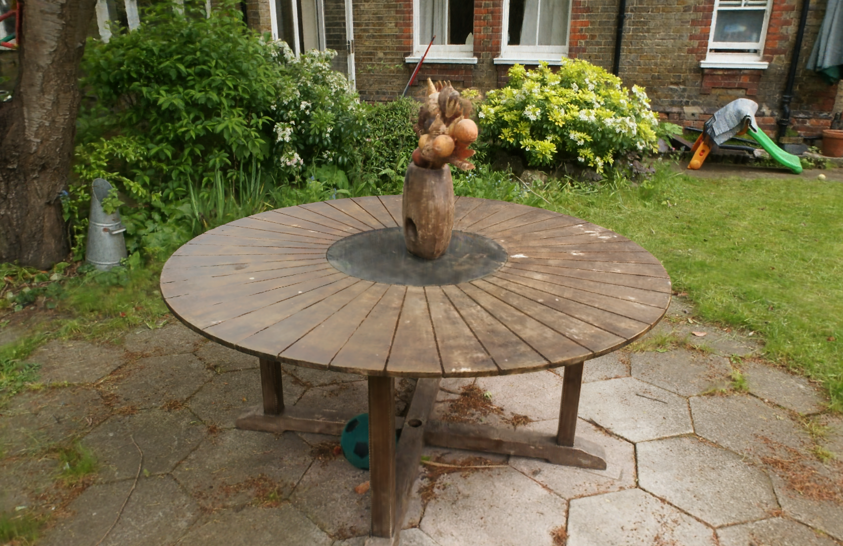}{1.4,1.4}{2.5,0.5} \\

		\spyimage{\resultfigwidth}{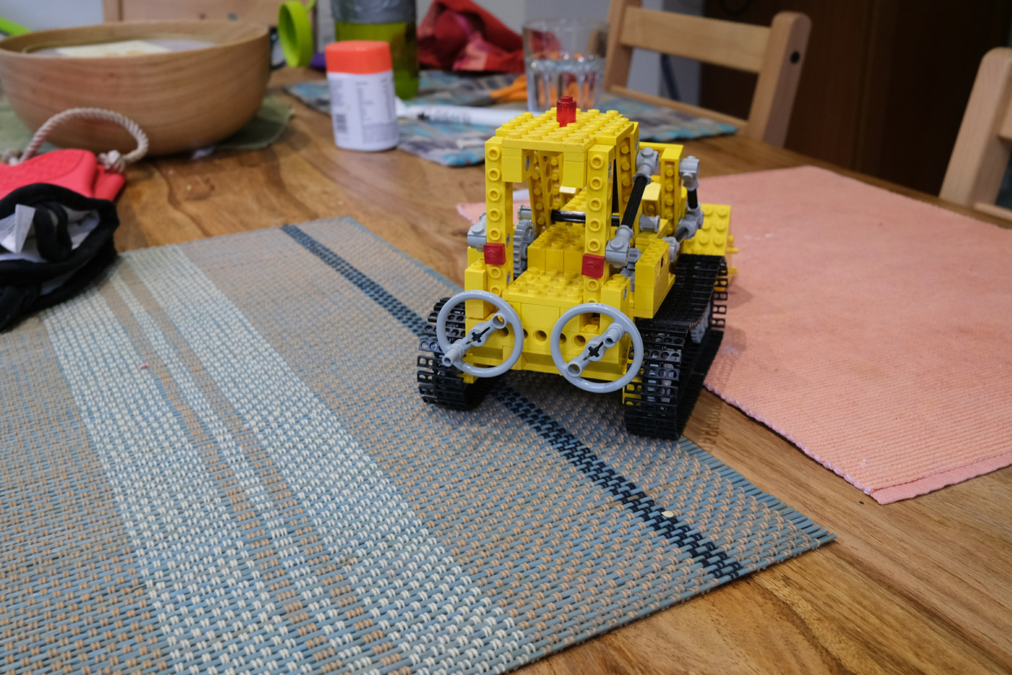}{1.2,0.915}{1,1} &
		\spyimage{\resultfigwidth}{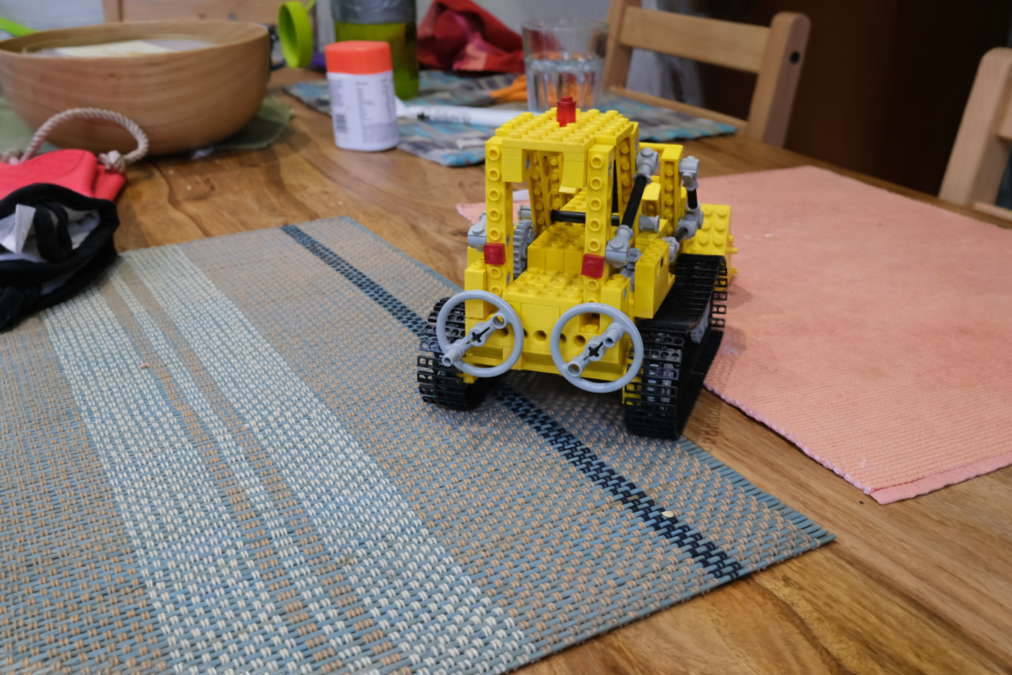}{1.2,0.915}{1,1} &
		\spyimage{\resultfigwidth}{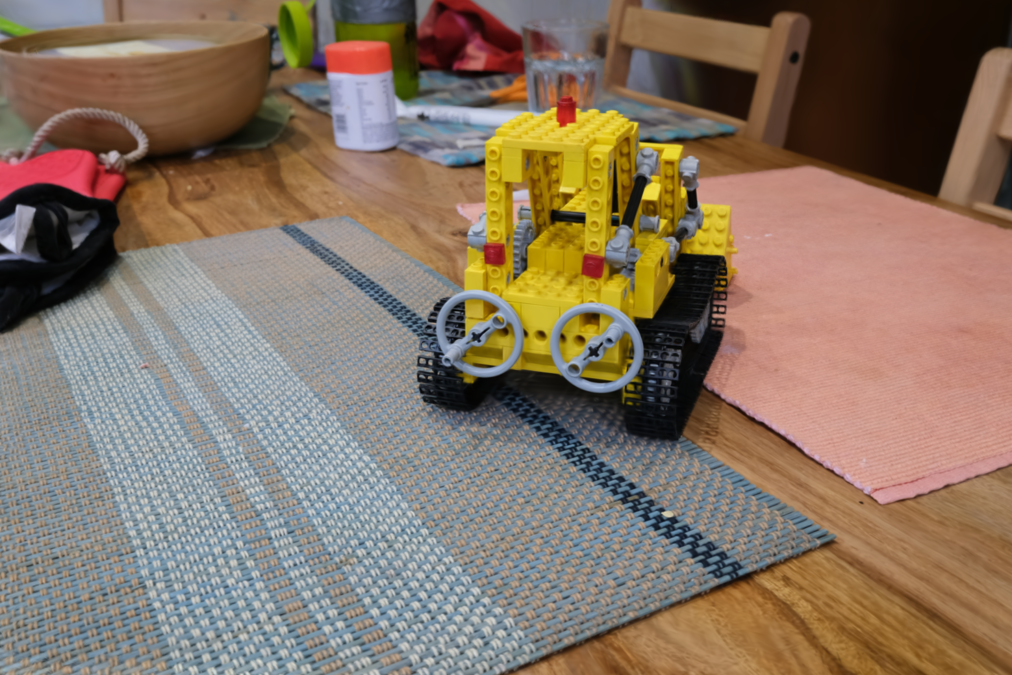}{1.2,0.915}{1,1} &
		\spyimage{\resultfigwidth}{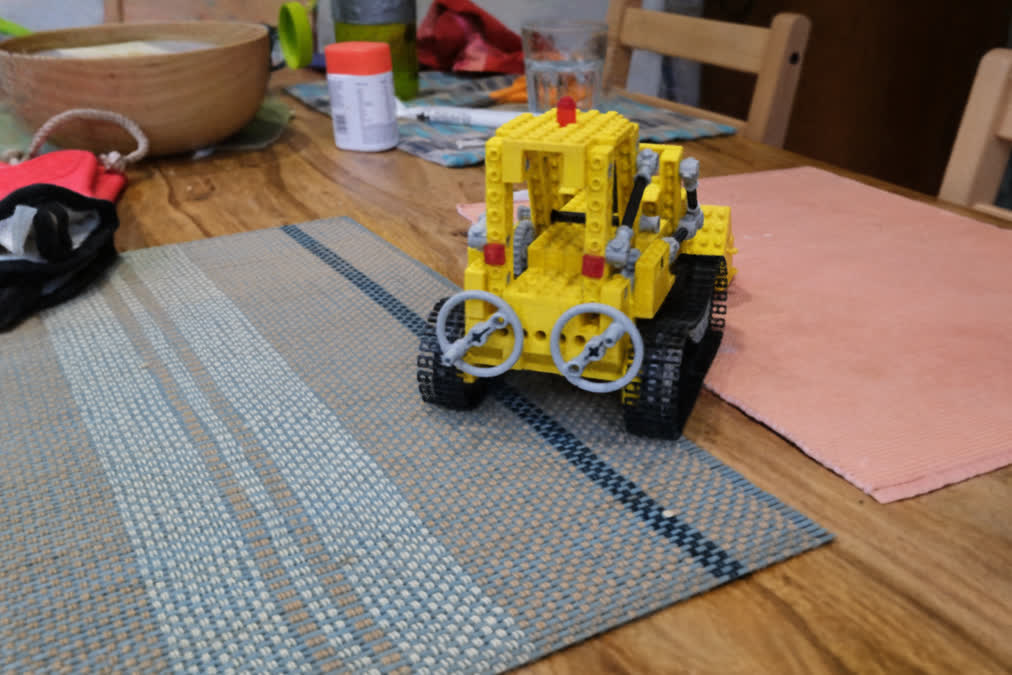}{1.2,0.915}{1,1} &
		\spyimage{\resultfigwidth}{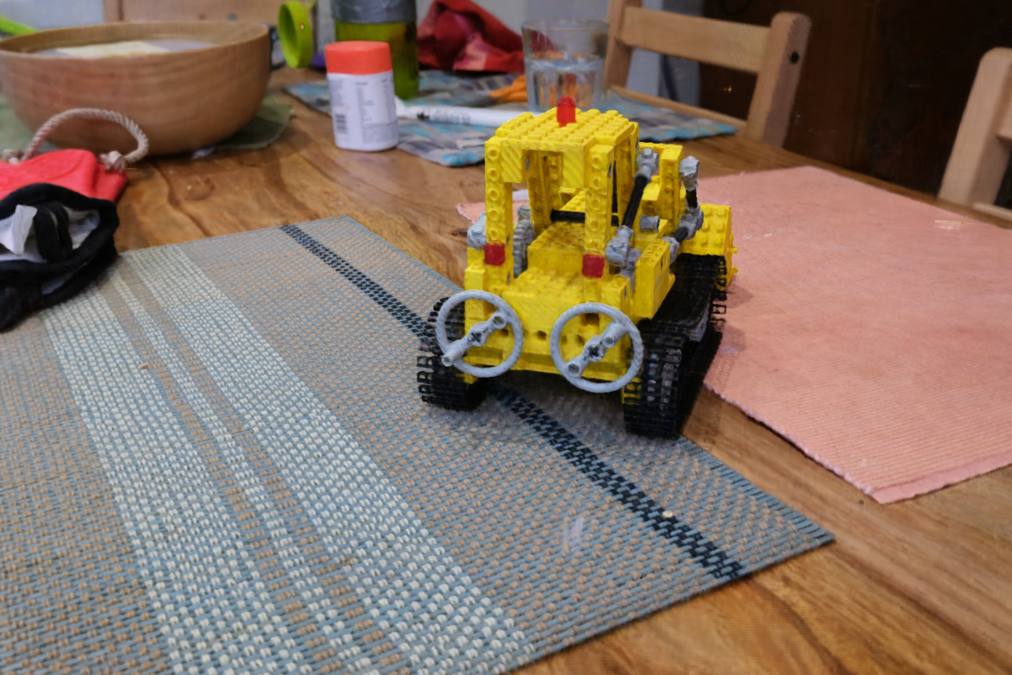}{1.2,0.915}{1,1}
	\end{tabular}
	\caption{
		\label{fig:res2}
		Visual comparison between INGP, MeRF, 3DGS and ours.
	}
\end{figure}

We also show screenshots from our WebGL application on a phone in 
the supplemental video.
The initial loading time for the full 3DGS scene on our local wifi network was 120sec, displaying at 16FPS, while using our method, the download time is only 5sec and 45FPS; a $\times24$ speedup in download time. 

\begin{table*}[!h]
	\caption{
		\label{tab:comparisons} {
			Quantitative comparison of relevant methods; the first part of the table is taken from Table 1 of the 3DGS publication. Results are computed over three representative datasets. Our method achieves the best compromise between size, speed, and quality. M-NeRF360 is consistently the smallest but trains for days and needs minutes to render, while INGP has significantly lower quality and rendering speed, for only a marginal gain in storage. Red is the best method, then orange, then yellow; the same color code is used for all tables.
	}}
	\small
	\scalebox{0.8}{	
	\begin{tabular}{l|m{0.6cm}m{0.6cm}m{0.6cm}cm{1.3cm}m{0.8cm}|m{0.6cm}m{0.6cm}m{0.6cm}cm{1.2cm}m{0.8cm}}
			
			Dataset & \multicolumn{6}{c|}{Mip-NeRF360}  & \multicolumn{6}{c}{Tanks\&Temples} \\
			Method/Metric
			& $SSIM^\uparrow$   & $PSNR^\uparrow$    & $LPIPS^\downarrow$  & Train & FPS~(Time) & Mem 
			& $SSIM^\uparrow$   & $PSNR^\uparrow$    & $LPIPS^\downarrow$  & Train & FPS~(Time) & Mem \\
			\hline 
			Plenoxels	& 0.626 & 23.08 & 0.463 & 25m49s & 6.79~(147ms) & 2.1GB 					& 0.719 & 21.08 & 0.379 & 25m5s  & 13.0~(77ms) & 2.3GB 						\\
			INGP		& 0.671 & 25.30 & 0.371 & 5m37s	 & 11.7~(85ms)  & \cellcolor{orange!40}13MB & 0.723 & 21.72 & 0.330 & 5m26s  & 17.1~(58ms) & \cellcolor{yellow!40}13MB 	\\ 
			M-NeRF360	& 0.792 & 27.69 & 0.237 & 48h 	 & 0.06~(16.7s) & \cellcolor{red!40}8.6MB  	& 0.759 & 22.22 & 0.257 & 48h 	 & 0.14~(7.1s)& \cellcolor{red!40}8.6MB 	    \\
			MeRF		& 0.722 & 25.24 & 0.311 & - 	 & 162~(6.1ms)  & 162MB 					&   - 	&   - 	&   - 	&   - 	 & -    & - 							\\
			3DGS		& 0.815 & 27.21 & 0.214 & 41m33s & 134~(7.5ms)  & 734MB 					& 0.841 & 23.14 & 0.183 & 26m54s & 154~(6.5ms) & 411MB 						\\
			3DGS*       & 0.813 & 27.42 & 0.217 & 31m16s & 178~(5.6ms)  & 744MB 					& 0.844 & 23.66 & 0.178 & 17m47s & 227~(4.4ms) & 412MB  					    \\
			Ours        & 0.809 & 27.10 & 0.226 & 25m27s & 284~(3.5ms)  &  29MB 					& 0.840 & 23.57 & 0.188 & 14m0s  & 433~(2.3ms) &  14MB 						\\
			Low  		& 0.811 & 27.22 & 0.224 & 25m22s & 295~(3.4ms)  &  46MB 					& 0.841 & 23.64 & 0.186 & 14m4s  & 436~(2.3ms) &  21MB 						\\
			High 		& 0.806 & 27.02 & 0.230 & 25m7s  & 298~(3.3ms)  & \cellcolor{yellow!40}23MB & 0.836 & 23.28 & 0.192 & 13m41s & 468~(2.1ms) &  \cellcolor{orange!40}10MB    \\
			
		\end{tabular}
	}
	\newline
	\vspace*{0.3cm}
	\newline
		\scalebox{0.8}{
			
		\begin{tabular}{l|m{0.6cm}m{0.6cm}m{0.6cm}cm{1.3cm}m{0.8cm}}
			
			 & \multicolumn{6}{c}{Deep Blending}\\
			Method/Metric
			& $SSIM^\uparrow$   & $PSNR^\uparrow$    & $LPIPS^\downarrow$  & Train & FPS~(Time) & Mem \\
			\hline 
			Plenoxels	& 0.795 & 23.06 & 0.510 & 27m49s & 11.2~(89ms) & 2.7GB \\
			INGP		& 0.797 & 23.62 & 0.423 & 6m31s & 3.26~(307ms) & \cellcolor{orange!40}13MB \\ 
			M-NeRF360	& 0.901 & 29.40 & 0.245 & 48h & 0.09~(11.1s) & \cellcolor{red!40}8.6MB\\
			MeRF		& - & - & - & - & - & - \\
			3DGS		& 0.903 & 29.41 & 0.243 & 36m2s & 137~(7.3ms) & 676MB\\
			3DGS*       & 0.899 & 29.47 & 0.247 & 28m2s & 201~(5ms) & 630MB \\
			Ours        & 0.902 & 29.63 & 0.249 & 22m4s & 360~(2.8ms) &  \cellcolor{yellow!40}18MB \\
			Low  		& 0.903 & 29.74 & 0.248 & 21m59s & 371~(2.7ms) &  35MB \\
			High 		& 0.902 & 29.56 & 0.251 & 21m31s & 406~(2.5ms) &  \cellcolor{orange!40}13MB \\
		\end{tabular}
	}
\end{table*}

\subsection{Evaluation}
\label{sec:eval}
We compare our proposed method to previous solutions. In Tab.~\ref{tab:comparisons}, we reproduce the results from the original 3DGS paper and have added four rows: one for the recently published MeRF method, one for 3DGS* corresponding to the runs with the public codebase (see previous footnote) and Our full solution, as well as the low- and high-compression variants. We show results for 3DGS-based methods after 30K training iterations.
To illustrate the possible tradeoffs for different target use cases, we also include two variants with slightly different configurations, achieving different degrees of compression: "Low" and "High". These variants use parameters $\epsilon_\sigma =$ 0.01, $\epsilon_{cdist} =$ 0.0068 and $\epsilon_\sigma =$ 0.06, $\epsilon_{cdist} =$ 0.054, respectively. Furthermore, the minimum redundancy score to be considered for culling is set to 2 for the high-compression variant.
We see that our proposed method and its variants are competitive in memory compared to INGP and even MipNerf360, but maintain the advantages of 3DGS w.r.t.\ speed and quality, including for training time.

We next perform a set of ablation studies to analyze the effect of our choices on memory and quality.
We investigate the effect of various parameters, in particular culling based solely on opacity vs. using our redundancy metric and selecting high-scoring candidates either with low opacity or at random.
The results are summarized in Tab.~\ref{tab:ablations}. While both opacity culling and redundancy reduction have comparable effects, they identify and cull \emph{different} sets of primitives. By combining the two policies, we get the highest reduction while maintaining robust quality metrics.
Tab.~\ref{tab:quanthalf-ablation} quantifies the impact of our quantization from full 32-bit floats to half floats. Again, the impact is minor when compared to the significant reduction in required file size.

\begin{table*}[!t]
	\caption{
		\label{tab:ablations}{Ablation on primitive culling approaches. We evaluate the effectiveness of our culling strategy against simpler baselines. First, we evaluate a straightforward culling approach that culls points with the lowest opacity across the scene (Opacity). Second, we compute candidates for pruning based on our spatial redundancy score (see Sec.~\ref{sec:point-removal}) \emph{and} whose opacity is in the lowest 50\% (Redundancy). Third, we compute the same candidates but prune the 50\% of the points at random (Redundancy Random) and finally, we evaluate our full method that combines Opacity (1st row) and Redundancy (2nd row). On average, the combination of the two methods achieves ~9\% higher reduction of points than the best-performing other method.}
	}
	\small
	\scalebox{0.78}{
		\begin{tabular}{l|rrr|rrr|rrr}
			Dataset & \multicolumn{3}{c|}{Mip-NeRF360} & \multicolumn{3}{c|}{Tanks\&Temples} & \multicolumn{3}{c}{Deep Blending} \\
			Method/Metric &        $PSNR^\uparrow$ & \multicolumn{2}{c|}{\#Prim (\%Base)} &          $PSNR^\uparrow$ & \multicolumn{2}{c|}{\#Prim (\%Base)} &          $PSNR^\uparrow$ & \multicolumn{2}{c}{\#Prim (\%Base)} \\
			\midrule
			Opacity           &       27.16 &        1.64M & (48\%) &         23.52 &        0.93M & (50\%) &         29.59 &        1.37M & (48\%) \\
			Redundancy        &       27.16 &        1.88M & (56\%) &         23.51 &        0.85M & (50\%) &         29.57 &        1.30M & (45\%) \\
			Redundancy Random &       27.16 &        1.91M & (57\%) &         23.55 &        0.87M & (51\%) &         29.63 &        1.33M & (46\%) \\
			Ours           &       27.10 &        1.46M & (43\%) &         23.57 &        0.68M & (39\%) &         29.63 &        1.01M & (35\%) \\
		\end{tabular}
	}
\end{table*}

\begin{table*}[!t]
	\caption{
		\label{tab:quanthalf-ablation} {
			We show the effect of half-precision floats on the PSNR, Memory, and Position memory footprint fraction of total model memory.
			The drop in quality is limited to 0.07 dB, while the memory reduction is increased by at least 20\%
	}}
	\small
	\scalebox{0.78}{
		\begin{tabular}{l|rrrr|rrrr|rrrr}
			Dataset & \multicolumn{4}{c|}{Mip-NeRF360} & \multicolumn{4}{c|}{Tanks\&Temples} & \multicolumn{4}{c}{Deep Blending} \\
			Method/Metric &       $PSNR^\uparrow$ & \multicolumn{2}{c}{Mem ($\times$ Gain)} & Pos \% &          $PSNR^\uparrow$ & \multicolumn{2}{c}{Mem ($\times$ Gain)} & Pos \% &          $PSNR^\uparrow$ & \multicolumn{2}{c}{Mem ($\times$ Gain)} & Pos \% \\
			\midrule
			Baseline     &       27.42 &        744MB &  ($\times$1.0) &    5\% &         23.66 &        412MB &  ($\times$1.0) &    5\% &         29.47 &        630MB &  ($\times$1.0) &    5\% \\
			K-means      &       27.17 &         38MB & ($\times$20.0) &   43\% &         23.60 &         18MB & ($\times$21.7) &   41\% &         29.65 &         24MB & ($\times$26.5) &   47\% \\
			+Half Floats &       27.10 &         29MB & ($\times$25.7) &   28\% &         23.57 &         14MB & ($\times$27.6) &   26\% &         29.63 &         18MB & ($\times$34.8) &   31\% \\
		\end{tabular}
	}
\end{table*}

In the Appendix, we provide additional comparisons with concurrent, unpublished work that also address compact 3DGS representation. 
Comparing with reported metrics for the (as of yet unpublished) techniques, we find that our method provides a superior quality/storage tradeoff.
We achieve the highest compression rates across all datasets with negligible differences in quality. We often achieve higher compression rates and higher PSNR scores than the best competing methods at the same time.
This is due to the fact that our compression uses both an effective quantization and an effective pruning technique of unused elements in its representation.

\section{Conclusion}

In this paper, we have presented a complete and efficient memory reduction method for 3DGS. 
We achieve this by introducing a resolution-aware primitive reduction method, reducing the number of primitives by half, an adaptive adjustment method to choose the appropriate number of SH bands required for each primitive, and a codebook-based quantization method.

Our method results in a $\times$27 reduction in memory with a $\times$1.7 increase in rendering speed. We demonstrate our results in the context of a streaming setup with a WebGL implementation for rendering, reducing download time 20-30 times and increasing rendering speed approximately 3 times. Ours is the first such streaming/mobile 3DGS solution that preserves high visual quality.
Our memory reduction removes one significant limitation of 3DGS; Ours is thus the most competitive NVS method for all three criteria: speed, quality, and memory consumption.

In future work, it would be interesting to investigate how to further reduce the number of primitives required, and more importantly \emph{avoid} the over-densification in the first place. Simple initial tests have shown that this is a very hard problem; one possible direction would be the use of data-driven priors, for example, supervision on depth~\cite{chung2023depth}.

\section{Acknowledgments}
This research was funded by the ERC Advanced grant FUNGRAPH No 788065 (\href{https://fungraph.inria.fr}{https://fungraph.inria.fr}). The authors are grateful to Adobe for generous donations, NVIDIA for a hardware donation, and the OPAL infrastructure from Universit\'e C\^ote d'Azur. 

\section*{Appendix}

In this section, we compare our full method and our low compression variant to several concurrent methods that are unpublished preprints in Tab.~\ref{tab:compare-arxiv} and \ref{tab:compare-arxiv2}.
We report two tables to ensure that all results are presented using exactly the same datasets.
The authors of Compact3D \cite{navaneet2023compact3d} only report an average size across the three datasets of 54MB.
For comparison, the relevant number for our full method and low compression variant is 26MB and 41MB, respectively.
While Compressed3D \cite{westermann} achieves impressive compression rates on disk, these do not directly map to VRAM consumption during rendering, since their use of the \texttt{DEFLATE} algorithms accounts for a factor of $\approx2\times$.

\begin{table*}[!ht]
	\caption{
		\label{tab:compare-arxiv}
		Comparisons to unpublished concurrent methods (preprints). The results shown here are those that include the full set of Mip-NeRF360 scenes, i.e., the two scenes with licensing issues, \texttt{treehill} and \texttt{flowers}.
	}
    \small
	\scalebox{0.78}{
		\begin{tabular}{l|m{0.65cm}m{0.65cm}m{0.65cm}rc|m{0.65cm}m{0.65cm}m{0.65cm}rc}
			Dataset & \multicolumn{5}{c|}{Mip-NeRF360} & \multicolumn{5}{c}{Deep Blending} \\
			Method/Metric &        $SSIM^\uparrow$ &  $PSNR^\uparrow$ & $LPIPS^\downarrow$ &  Train &  Mem &          $SSIM^\uparrow$ &  $PSNR^\uparrow$ & $LPIPS^\downarrow$ &  Train & Mem \\
			\midrule
			Ours                & \cellcolor{orange!40}0.809 & \cellcolor{yellow!40}27.10 & \cellcolor{orange!40}0.226 & 25m27s & \cellcolor{red!40}29MB &         \cellcolor{yellow!40}0.902 & 29.63 & 0.249 &    22m4s & \cellcolor{red!40}18MB \\
			Low                 & \cellcolor{red!40}0.811 & \cellcolor{red!40}27.22 & \cellcolor{red!40}0.224 & 25m22s & \cellcolor{orange!40}46MB &         \cellcolor{orange!40}0.903 & 29.74 & \cellcolor{yellow!40}0.248 &  21m59s & \cellcolor{yellow!40}35MB \\
			EAGLES~\cite{girish2023eagles}              & \cellcolor{yellow!40}0.808 & \cellcolor{orange!40}27.16 & 0.238 & 19m57s & 68MB &         \cellcolor{red!40}0.910 & \cellcolor{red!40}29.91 & \cellcolor{red!40}0.245 & 17m24s & 62MB \\
			Compact3D~\cite{navaneet2023compact3d}           & \cellcolor{yellow!40}0.808 & \cellcolor{orange!40}27.16 & \cellcolor{yellow!40}0.228 &   -     & - &         \cellcolor{orange!40}0.903 & \cellcolor{yellow!40}29.75 & \cellcolor{orange!40}0.247 &   -     & - \\
			Compressed3D~\cite{westermann} & 0.801 & 26.98 & 0.238 & - & \cellcolor{red!40}29MB &  0.898 & 29.38 & 0.253 & - & \cellcolor{orange!40}25MB\\
			Compact3DGS~\cite{lee2023compact} & 0.798 & 27.08 & 0.247 &  33m6s & \cellcolor{yellow!40}48MB &         0.901 & \cellcolor{orange!40}29.79 & 0.258 & 27m33s & 43MB \\

		\end{tabular}
	}
\end{table*}

\begin{table*}[!ht]
	\caption{
		\label{tab:compare-arxiv2}
		Comparisons to unpublished concurrent methods (preprints). The results shown here exclude the two scenes with licensing issues, \texttt{treehill} and \texttt{flowers}. 
	}
    \small
	\scalebox{0.78}{
		\begin{tabular}{l|m{0.65cm}m{0.65cm}m{0.65cm}rc|m{0.65cm}m{0.65cm}m{0.65cm}rc}
			Dataset & \multicolumn{5}{c|}{Mip-NeRF360 No Hidden} & \multicolumn{5}{c}{Tanks\&Temples} \\
			Method/Metric &                  $SSIM^\uparrow$ &  $PSNR^\uparrow$ & $LPIPS^\downarrow$ &  Train &  Mem &          $SSIM^\uparrow$ &  $PSNR^\uparrow$ & $LPIPS^\downarrow$ &  Train &  Mem \\
			\midrule
			Ours                &            \cellcolor{orange!40}0.864 & 28.58 &\cellcolor{orange!40} 0.193 &   26m0s & \cellcolor{red!40}27MB &         \cellcolor{orange!40}0.840 & \cellcolor{orange!40}23.57 & \cellcolor{orange!40}0.188 & 14m0s & \cellcolor{red!40}14MB \\
			Low                 &             \cellcolor{red!40}0.866 &  \cellcolor{red!40}28.73 &  \cellcolor{red!40}0.190 &  25m52s & \cellcolor{yellow!40}43MB &          \cellcolor{red!40}0.841 &  \cellcolor{red!40}23.64 &  \cellcolor{red!40}0.186 &  14m4s & \cellcolor{yellow!40}21MB \\
			EAGLES~\cite{girish2023eagles}              &             \cellcolor{red!40}0.866 & \cellcolor{orange!40}28.69 & \cellcolor{yellow!40}0.200 & 20m18s & 67MB &         \cellcolor{yellow!40}0.835 & 23.41 & \cellcolor{yellow!40}0.200 &  9m48s & 34MB \\
			Compact3D~\cite{navaneet2023compact3d}           &              -    &  -     & -      & -   &    -  &         \cellcolor{orange!40}0.840 & \cellcolor{yellow!40}23.47 & \cellcolor{orange!40}0.188 & -    & - \\
			Compressed3D~\cite{westermann} & 0.857 & 28.48 & 0.205 & - & \cellcolor{orange!40}28MB & 0.832 & 23.32 & 0.194 & - & \cellcolor{orange!40}17MB\\
			Compact3DGS~\cite{lee2023compact}        &            0.856 & \cellcolor{yellow!40}28.60 & 0.209 &  33m1s & 46MB &         0.832 & 23.31 & 0.202 & 18m19s & 39MB \\
			LightGaussian~\cite{fan2023lightgaussian}       &            \cellcolor{yellow!40}0.858 & 28.46 & 0.210 &   -     & 42MB &         0.807 & 22.83 & 0.242 &   -     & 22MB \\
		\end{tabular}
	}
\end{table*}

\begin{table*}[!ht]
    \caption{
        \label{tab:per-scene-m360}
        Per-scene results for Mip-NeRF 360 dataset.
        }
        \small
	\scalebox{0.78}{
        \begin{tabular}{l|m{0.55cm}m{0.55cm}m{0.55cm}rm{0.6cm}m{0.8cm}m{1.3cm}|m{0.55cm}m{0.55cm}m{0.55cm}rm{0.6cm}m{0.6cm}m{1.3cm}}
            & \multicolumn{7}{c|}{Baseline} & \multicolumn{7}{c}{Ours} \\
            Scene    &$SSIM^\uparrow$   &  $PSNR^\uparrow$ & $LPIPS^\downarrow$ &  Train &  \#Prim &    Mem & FPS~(Time) & $SSIM^\uparrow$ &  $PSNR^\uparrow$ & $LPIPS^\downarrow$ &  Train &  \#Prim &    Mem & FPS~(Time) \\
            \midrule
            bicycle  & 0.763 & 25.10 & 0.212 & 41m18s & 6.05M & 1362MB & 115~(8.7ms) & 0.761 & 25.06 & 0.221 &  30m2s & 2.41M & 48MB & 260~(3.8ms) \\ 
            flowers  & 0.604 & 21.44 & 0.338 & 28m39s & 3.63M &  816MB & 205~(4.9ms) & 0.598 & 21.44 & 0.346 &  23m6s & 1.61M & 35MB & 353~(2.8ms) \\ 
            garden   & 0.863 & 27.28 & 0.108 &  43m2s & 5.77M & 1298MB & 127~(7.9ms) & 0.854 & 27.03 & 0.119 & 31m52s & 2.36M & 47MB & 245~(4.1ms) \\ 
            stump    & 0.771 & 26.60 & 0.216 & 33m12s & 4.75M & 1068MB & 216~(4.6ms) & 0.776 & 26.68 & 0.219 & 26m58s & 2.50M & 48MB & 321~(3.1ms) \\ 
            treehill & 0.633 & 22.44 & 0.327 &  29m4s & 3.80M &  854MB & 192~(5.2ms) & 0.631 & 22.44 & 0.337 & 23m55s & 1.81M & 37MB & 286~(3.5ms) \\ 
            room     & 0.917 & 31.44 & 0.221 & 26m14s & 1.53M &  345MB & 179~(5.6ms) & 0.913 & 30.95 & 0.228 & 22m38s & 0.56M & 10MB & 272~(3.7ms) \\ 
            counter  & 0.906 & 28.98 & 0.202 & 25m32s & 1.20M &  269MB & 174~(5.7ms) & 0.898 & 28.54 & 0.213 & 22m57s & 0.50M & 11MB & 256~(3.9ms) \\ 
            kitchen  & 0.925 & 31.27 & 0.127 & 31m32s & 1.79M &  403MB & 145~(6.9ms) & 0.917 & 30.52 & 0.136 & 27m20s & 0.84M & 18MB & 224~(4.5ms) \\ 
            bonsai   & 0.940 & 32.21 & 0.206 & 22m53s & 1.25M &  281MB & 254~(3.9ms) & 0.932 & 31.26 & 0.216 & 20m18s & 0.52M & 10MB & 341~(2.9ms) \\ 
        \end{tabular}
    }
\end{table*}

\begin{table*}[!ht]
    \caption{
        \label{tab:per-scene-tat}
        Per-scene results for Tanks\&Temples dataset.
        }
        \small
	\scalebox{0.78}{
            \begin{tabular}{l|m{0.55cm}m{0.55cm}m{0.55cm}rm{0.6cm}m{0.7cm}m{1.3cm}|m{0.55cm}m{0.55cm}m{0.55cm}rm{0.6cm}m{0.6cm}m{1.3cm}}
                     & \multicolumn{7}{c|}{Baseline} & \multicolumn{7}{c}{Ours} \\
            Scene    &$SSIM^\uparrow$   &  $PSNR^\uparrow$ & $LPIPS^\downarrow$ &  Train &  \#Prim &    Mem & FPS~(Time) & $SSIM^\uparrow$ &  $PSNR^\uparrow$ & $LPIPS^\downarrow$ &  Train &  \#Prim &    Mem & FPS~(Time) \\
            \midrule
            truck &    0.878 & 25.41 & 0.148 & 21m19s & 2.58M & 580MB & 214~(4.7ms) &    0.874 & 25.22 & 0.155 & 16m11s & 0.85M & 16MB & 418~(2.4ms) \\
            train &    0.810 & 21.91 & 0.208 & 14m16s & 1.08M & 243MB & 240~(4.2ms) &    0.805 & 21.93 & 0.222 & 11m50s & 0.50M & 12MB & 449~(2.2ms) \\
        \end{tabular}
    }
\end{table*}

\begin{table*}[!ht]
    \caption{
        \label{tab:per-scene-db}
        Per-scene results for DeepBlending dataset.
        }
        \small
	\scalebox{0.78}{
            \begin{tabular}{l|m{0.55cm}m{0.55cm}m{0.55cm}rm{0.6cm}m{0.7cm}m{1.3cm}|m{0.55cm}m{0.55cm}m{0.55cm}rm{0.6cm}m{0.6cm}m{1.3cm}}
                     & \multicolumn{7}{c|}{Baseline} & \multicolumn{7}{c}{Ours} \\
            Scene    &$SSIM^\uparrow$   &  $PSNR^\uparrow$ & $LPIPS^\downarrow$ &  Train &  \#Prim &    Mem & FPS~(Time) & $SSIM^\uparrow$ &  $PSNR^\uparrow$ & $LPIPS^\downarrow$ &  Train &  \#Prim &    Mem & FPS~(Time) \\
            \midrule
            drjohnson &    0.898 & 29.03 & 0.247 & 31m22s & 3.27M & 736MB & 171~(5.8ms) &    0.901 & 29.21 & 0.249 & 24m38s & 1.25M & 23MB & 335~(3ms) \\
            playroom  &    0.900 & 29.90 & 0.247 & 24m43s & 2.33M & 523MB & 232~(4.3ms) &    0.904 & 30.05 & 0.249 & 19m30s & 0.77M & 13MB & 385~(2.6ms) \\
        \end{tabular}
    }
\end{table*}

We find that our low-compression variant already yields smaller file sizes than the most compact competitors, but maintains image quality  closest to 3DGS across the board. We note that for Deep Blending, more invasive regularization in some other methods can \emph{improve} the metrics over 3DGS. 
Our proposed variant still achieves competitive quality, while requiring even less storage, resulting in significantly smaller files and an extremely favorable file size/quality tradeoff.

Finally,
we provide per-scene results for Mip-NeRF360 Tab.~\ref{tab:per-scene-m360},
Tanks\&Temples Tab.~\ref{tab:per-scene-tat}
and DeepBlending Tab.~\ref{tab:per-scene-db} datasets.

\bibliographystyle{ACM-Reference-Format}
\bibliography{sample-base}

\end{document}